\documentclass{article}
\pdfoutput=1

\usepackage{times}
\usepackage{graphicx}
\usepackage{subfigure} 
\usepackage{amssymb,amsmath}
\usepackage{mathtools}
\mathtoolsset{showonlyrefs}
\usepackage{natbib}
\usepackage{algorithm}
\usepackage{algorithmic}
\usepackage{hyperref}
\usepackage{enumitem}

\newtheorem{mytheorem}{Theorem} 
\newtheorem{mylemma}{Lemma}

\usepackage[accepted]{icml2016}

\def \mytitlelong {Polynomial Networks and Factorization
Machines:\\ New Insights and Efficient Training Algorithms}
\def \mytitle {Polynomial Networks and Factorization
Machines: New Insights and Efficient Training Algorithms}

\icmltitlerunning{\mytitle}

\DeclareMathOperator*{\argmin}{argmin}

\DeclareMathOperator*{\rank}{rank}

\DeclareMathOperator*{\sign}{sign}
\DeclareMathOperator*{\diag}{diag}

\newcommand\bs[1]{\boldsymbol{#1}}
\newcommand\col[1]{\boldsymbol{\bar{#1}}}
\newcommand\innerprod[2]{\langle \bs{#1}, \bs{#2} \rangle}

\newcommand\partialfrac[2]{\frac{\partial #1}{\partial #2}}
\newcommand\tensor[1]{\boldsymbol{\mathcal{#1}}}

\newcommand\tr{\mathrm{T}}
\newcommand\qed{\square}

\newcommand\symtensor[2]{#1^{\otimes #2}}

\newcommand\eigen[2]{\bs{#1}\diag(\bs{#2})\bs{#1}^\tr}

\newcommand\decomp[2]{\bs{#1}\bs{#2}^\tr}
\newcommand\frobnorm[1]{\|\bs{#1}\|_F}
\newcommand\symmat[1]{\frac{1}{2}(\bs{#1} + \bs{#1}^\tr)}
\newcommand\sumfrobnorm[2]{(\frobnorm{#1}^2 + \frobnorm{#2}^2)}
\newcommand\symop[1]{\mathcal{S}(#1)}

\newcommand\polykern[4]{\mathcal{P}^{#3}_{#4}(\bs{#1}, \bs{#2})}

\newcommand\homogkern[3]{\mathcal{H}^{#3}(\bs{#1}, \bs{#2})}

\newcommand\combikern[3]{\mathcal{A}^{#3}(\bs{#1}, \bs{#2})}

\newcommand\diagkern[3]{\mathcal{D}^{{#3}}(\bs{#1}, \bs{#2})}
\newcommand\negvec[2]{\bs{#1}_{\neg #2}}

\def\multisetcoef#1#2{\ensuremath{\left(\kern-.3em\left(\genfrac{}{}{0pt}{}{#1}{#2}\right)\kern-.3em\right)}}

\begin{document} 

\twocolumn[
\icmltitle{\mytitlelong}

\icmlauthor{Mathieu Blondel}{mathieu.blondel@lab.ntt.co.jp}
\icmlauthor{Masakazu Ishihata }{ishihata.masakazu@lab.ntt.co.jp}
\icmlauthor{Akinori Fujino}{fujino.akinori@lab.ntt.co.jp}
\icmlauthor{Naonori Ueda}{ueda.naonori@lab.ntt.co.jp}
\icmladdress{NTT Communication Science Laboratories,
2-4 Hikaridai Seika-cho Soraku-gun, Kyoto 619-0237, Japan}

\icmlkeywords{polynomial kernel, low-rank, factorization machines}

\vskip 0.3in
]


\begin{abstract} 
Polynomial networks and factorization machines are two recently-proposed models
that can efficiently use feature interactions in classification and regression
tasks.  In this paper, we revisit both models from a unified perspective.  Based
on this new view, we study the properties of both models and propose new
efficient training algorithms.  Key to our approach is to cast parameter
learning as a low-rank symmetric tensor estimation problem, which we solve by
multi-convex optimization.  We demonstrate our approach on regression and
recommender system tasks.
\end{abstract} 

\vspace{-0.8cm}
\section{Introduction}

Interactions between features play an important role in many classification and
regression tasks.  One of the simplest approach to leverage such interactions
consists in \textit{explicitly} augmenting feature vectors with products of
features (monomials), as in polynomial regression.  Although fast linear model
solvers can be used \citep{low_poly,coffin}, an obvious drawback of this kind of
approach is that the number of parameters to estimate scales as $O(d^m)$, where
$d$ is the number of features and $m$ is the order of interactions considered.
As a result, it is usually limited to second or third-order interactions.

Another popular approach consists in using a polynomial kernel so as to
\textit{implicitly} map the data via the kernel trick.  The main advantage of
this approach is that the number of parameters to estimate in the model is
actually independent of $d$ and $m$. However, the cost of storing and evaluating
the model is now proportional to the number of training instances. This is
sometimes called the curse of kernelization \citep{budgetpegasos}.  Common ways
to address the issue include
the Nystr\"{o}m method \citep{nystrom_method}, random
features \citep{dot_kernel} and sketching \citep{poly_fft,poly_embedding}.  

In this paper, in order to leverage feature interactions in possibly very
high-dimensional data, we consider models which predict the output $y \in
\mathbb{R}$ associated with an input vector $\bs{x} \in \mathbb{R}^d$ by
\begin{equation}
    \hat{y}_{\mathcal{K}}(\bs{x}; \bs{\lambda}, \bs{P}) \coloneqq \sum_{s=1}^k
    \lambda_s \mathcal{K}(\bs{p}_s, \bs{x}),
\label{eq:prediction_function}
\end{equation}
where $\bs{\lambda} = [\lambda_1, \dots, \lambda_k]^\tr \in \mathbb{R}^k$,
$\bs{P} = [\bs{p}_1, \dots, \bs{p}_k] \in \mathbb{R}^{d \times k}$,
$\mathcal{K}$ is a kernel and $k$ is a hyper-parameter. More specifically, we
focus on two specific choices of $\mathcal{K}$ which allow us to use feature
interactions: the homogeneous polynomial and the ANOVA kernels.  Our
contributions are as follows. We show (Section \ref{sec:kernels}) that choosing
one kernel or the other allows us to recover \textbf{polynomial networks} (PNs)
\citep{livni} and, surprisingly, \textbf{factorization machines} (FMs)
\citep{fm,libfm}.  Based on this new view, we show important properties of PNs
and FMs. Notably, we show for the first time that the objective function of
arbitrary-order FMs is multi-convex (Section \ref{sec:direct_optimization}).
Unfortunately, the objective function of PNs is not multi-convex. To remedy this
problem, we propose a \textit{lifted} approach, based on casting parameter
estimation as a low-rank tensor estimation problem (Section
\ref{sec:lifted_conversion}).  Combined with a symmetrization trick, this
approach leads to a multi-convex problem, for \textit{both} PNs and FMs (Section
\ref{sec:lifted_multi_convex}). 
We demonstrate our approach on regression and recommender system tasks.

\textbf{Notation.} We denote vectors, matrices and tensors using lower-case,
upper-case and calligraphic bold, e.g., $\bs{w}$, $\bs{W}$ and $\tensor{W}$.  We
denote the set of $\overbrace{d \times \dots \times d}^{m \text{ times}}$ real
tensors by $\mathbb{R}^{d^m}$ and the set of symmetric real tensors by
$\mathbb{S}^{d^m}$.  We use $\langle \cdot, \cdot \rangle$ to denote vector,
matrix and tensor inner product.  Given $\bs{x}$, we define a
symmetric rank-one tensor by $\symtensor{\bs{x}}{m} \coloneqq \bs{x} \otimes
\dots \otimes \bs{x} \in \mathbb{S}^{d^m}$, where
$(\symtensor{\bs{x}}{m})_{j_1,j_2,\dots,j_m} = x_{j_1} x_{j_2} \dots x_{j_m}$.
We use $[d]$ to denote the set $\{1,\dots,d\}$.

\section{Related work}

\subsection{Polynomial networks}

Polynomial networks (PNs) \citep{livni} of degree $m=2$ predict the output $y \in
\mathbb{R}$ associated with $\bs{x} \in \mathbb{R}^d$ by
\begin{equation}
    \hat{y}_{\text{PN}}(\bs{x};\bs{w}, \bs{\lambda}, \bs{P}) \coloneqq 
    \langle \bs{w}, \bs{x} \rangle +
    \langle \sigma(\bs{P}^\tr \bs{x}), \bs{\lambda} \rangle,
\label{eq:predict_pn}
\end{equation}
where $\bs{w} \in \mathbb{R}^d$, $\bs{P} \in \mathbb{R}^{d \times k}$,
$\bs{\lambda} \in \mathbb{R}^k$ and $\sigma(u) \coloneqq u^2$ is evaluated
element-wise. Intuitively, the right-hand term can be interpreted as a
feedforward neural network with one hidden layer of $k$ units and with
activation function $\sigma(u)$.  \citet{livni} also extend
\eqref{eq:predict_pn} to the case $m=3$ and show theoretically that PNs can
approximate feedforward networks with sigmoidal activation. A similar model was
independently shown to perform well on dependency parsing
\citep{dependency_parsing}. 
Unfortunately, the objective function of PNs is non-convex.  In Section
\ref{sec:lifted_optimization}, we derive a \textit{multi-convex} objective
based on low-rank symmetric tensor estimation, suitable for training
arbitrary-order PNs.

\subsection{Factorization machines}

One of the simplest way to leverage feature interactions is polynomial
regression (PR). For example, for second-order interactions, in this approach,
we compute predictions by
\begin{equation}
    \hat{y}_{\text{PR}}(\bs{x}; \bs{w}, \bs{W}) \coloneqq \langle \bs{w}, \bs{x}
    \rangle + \sum_{j' > j} \bs{W}_{j,j'} x_j x_{j'},
\end{equation}
where $\bs{w} \in \mathbb{R}^d$ and $\bs{W} \in \mathbb{R}^{d^2}$. Obviously,
model size in PR does not scale well w.r.t. $d$.  The main idea of (second-order)
factorization machines (FMs) \cite{fm,libfm} is to replace $\bs{W}$ with a
factorized matrix $\decomp{P}{P}$:
\begin{equation}
    \hat{y}_{\text{FM}}(\bs{x};\bs{w}, \bs{P}) \coloneqq \langle \bs{w}, \bs{x}
    \rangle + \sum_{j' > j} (\decomp{P}{P})_{jj'} x_j x_{j'},
\label{eq:predict_fm}
\end{equation}
where $\bs{P} \in \mathbb{R}^{d \times k}$. FMs have been increasingly popular
for efficiently modeling feature interactions in high-dimensional data, see
\citep{libfm} and references therein. In Section \ref{sec:direct_optimization},
we show for the first time that the objective function of arbitrary-order FMs is
\textit{multi-convex}.

\vspace{-0.2cm}
\section{Polynomial and ANOVA kernels}
\label{sec:kernels}

In this section, we show that the prediction functions used by polynomial
networks and factorization machines can be written using
\eqref{eq:prediction_function} for a specific choice of kernel.

The \textbf{polynomial kernel} is a popular kernel for using combinations
of features. The kernel is defined as
\begin{equation}
    \polykern{p}{x}{m}{\gamma} \coloneqq (\gamma +
    \innerprod{p}{x})^m,
\label{eq:inhomogeneous_kernel}
\end{equation}
\vspace{-0.1cm}
where $m \in \mathbb{N}$ is the degree and $\gamma > 0$ is a hyper-parameter.
We define the \textbf{homogeneous polynomial kernel} by
\begin{equation}
    \homogkern{p}{x}{m} \coloneqq  \polykern{p}{x}{m}{0} = \innerprod{p}{x}^m.
\label{eq:homogeneous_kernel}
\end{equation}
Let $\bs{p} = [p_1, \dots, p_d]^\tr$ and $\bs{x} = [x_1, \dots, x_d]^\tr$.
Then, 
\begin{equation} 
    \homogkern{p}{x}{m} = \sum_{j_1=1}^d \ldots \sum_{j_m=1}^d p_{j_1} x_{j_1}
    \ldots p_{j_m} x_{j_m}.  
\end{equation}
We thus see that $\mathcal{H}^m$ uses \textit{all} monomials of degree $m$
(i.e., all combinations of features \textit{with} replacement).

A much lesser known kernel is the \textbf{ANOVA kernel}
\citep{stitson,vapnik_book}. Following \citep[Section 9.2]{kernel_book}, the
ANOVA kernel of degree $m$, where $2 \le m \le d$, can be defined as
\begin{equation}
    \combikern{p}{x}{m} \coloneqq \sum_{j_m > \dots > j_1} p_{j_1} x_{j_1} \dots
    p_{j_m} x_{j_m}.
\label{eq:combination_kernel} 
\end{equation}
As a result, $\mathcal{A}^m$ uses only monomials composed of \textit{distinct}
features (i.e., feature combinations \textit{without} replacement). For
later convenience, we also define 
$\combikern{p}{x}{0} \coloneqq 1$ and
$\combikern{p}{x}{1} \coloneqq \innerprod{p}{x}$.

With $\mathcal{H}^m$ and $\mathcal{A}^m$
defined, we are now in position to state the following lemma.
\begin{mylemma}{Expressing PNs and FMs using kernels}

Let $\hat{y}_{\mathcal{K}}(\bs{x}; \bs{\lambda}, \bs{P})$ be defined as in
\eqref{eq:prediction_function}. Then,
\begin{align}
    \hat{y}_{\text{PN}}(\bs{x};\bs{w}, \bs{\lambda}, \bs{P}) &=
\langle \bs{w}, \bs{x} \rangle + 
\hat{y}_{\mathcal{H}^2}(\bs{x};\bs{\lambda},\bs{P}) \\
\hat{y}_{\text{FM}}(\bs{x};\bs{w}, \bs{P}) &=
\langle \bs{w}, \bs{x} \rangle + 
\hat{y}_{\mathcal{A}^2}(\bs{x};\bs{1},\bs{P}).
\end{align}
\end{mylemma}
\vspace{-0.2cm}
The relation easily extends to higher orders.
This new view allows us to state results that will be very useful in the next
sections. The first one is that $\mathcal{H}^m$ and $\mathcal{A}^m$ are
homogeneous functions, i.e., they satisfy
\begin{equation} 
    \lambda^m \mathcal{K}(\bs{p}, \bs{x}) = \mathcal{K}(\lambda \bs{p}, \bs{x})
\quad \mbox{$\forall \lambda \in \mathbb{R}, \forall m \in \mathbb{N}_+$}.
\label{eq:homogeneous_function}
\end{equation}
Another key property of $\combikern{p}{x}{m}$ is multi-linearity.\footnote{A
function $f(\theta_1, \dots, \theta_k)$ is called multi-linear (resp.
multi-convex) if it is linear (resp. convex) w.r.t. $\theta_1, \dots, \theta_k$
separately.}
\begin{mylemma}{Multi-linearity of $\combikern{p}{x}{m}$ w.r.t. $p_1,\dots,p_d$}

Let $\bs{p}, \bs{x} \in \mathbb{R}^d$, $j \in [d]$ and $1 \le m \le d$. Then,
\begin{equation} 
    \combikern{p}{x}{m} = \mathcal{A}^m(\negvec{p}{j}, \negvec{x}{j})
    + ~ p_j x_j ~ \mathcal{A}^{m-1}(\bs{p}_{\neg j}, \bs{x}_{\neg j}) 
\label{eq:recursion}
\end{equation}
where $\bs{p}_{\neg j}$ denotes the $(d-1)$-dimensional vector with $p_j$
removed and similarly for $\negvec{x}{j}$.
\label{lemma:multi_linearity}
\end{mylemma}
That is, everything else kept fixed, $\combikern{p}{x}{m}$ is an affine function
of $p_j$, $\forall j \in [d]$. Proof is given in Appendix
\ref{appendix:proof_multi_linearity}.

Assuming $\bs{p}$ is dense and $\bs{x}$ sparse, the cost of naively computing
$\combikern{p}{x}{m}$ by \eqref{eq:combination_kernel} is $O(n_z(\bs{x})^m)$,
where $n_z(\bs{x})$ is the number of non-zero features in $\bs{x}$.  To address
this issue, we will make use of the following lemma for computing
$\mathcal{A}^m$ in nearly $O(m n_z(\bs{x}))$ time when $m \in \{2,3\}$.
\begin{mylemma}{Efficient computation of ANOVA kernel}
\begin{equation}
\begin{aligned}
    \combikern{p}{x}{2} &= \frac{1}{2} &&\left[
\homogkern{p}{x}{2} -
\diagkern{p}{x}{2} \right] \\
\combikern{p}{x}{3} &= \frac{1}{6} &&\left[
\homogkern{p}{x}{3}  - 3
\diagkern{p}{x}{2,1} + 2 \diagkern{p}{x}{3} \right] \\
\end{aligned}
\end{equation}
where we defined
$\diagkern{p}{x}{m} \coloneqq \sum_{j=1}^d (p_j x_j)^m$ 
and
$\diagkern{p}{x}{m,n} \coloneqq \diagkern{p}{x}{m} \diagkern{p}{x}{n}$.
\label{lemma:linear_time_anova}
\end{mylemma}
See Appendix \ref{appendix:linear_time_imperative} for a
derivation.

\vspace{-0.2cm}
\section{Direct approach}
\label{sec:direct_optimization}

Let us denote the training set by $\bs{X} = [\bs{x}_1, \dots, \bs{x}_n] \in
\mathbb{R}^{d \times n}$  and $\bs{y} = [y_1, \dots, y_n]^\tr \in \mathbb{R}^n$.
The most natural approach to learn models of the form
\eqref{eq:prediction_function} is to directly choose $\bs{\lambda}$ and $\bs{P}$
so as to minimize some error function
\vspace{-0.12cm}
\begin{equation}
    D_{\mathcal{K}}(\bs{\lambda}, \bs{P}) \coloneqq
    \sum_{i=1}^n \ell\left(y_i, \hat{y}_{\mathcal{K}}(\bs{x}_i; \bs{\lambda},
    \bs{P}) \right),
\label{eq:direct_obj}
\end{equation}
where $\ell(y_i, \hat{y}_i)$ is a convex loss function.  Note that
\eqref{eq:direct_obj} is a convex objective w.r.t. $\bs{\lambda}$ regardless of
$\mathcal{K}$.
However, it is in general non-convex w.r.t.  $\bs{P}$.  Fortunately, when
$\mathcal{K} = \mathcal{A}^m$, we can show that \eqref{eq:direct_obj} is
multi-convex.
\vspace{-0.2cm}
\begin{mytheorem}{Multi-convexity of \eqref{eq:direct_obj} when $\mathcal{K} =
    \mathcal{A}^m$}
 
$D_{\mathcal{A}^m}$
    is convex in
$\bs{\lambda}$ and in each row of $\bs{P}$ separately.
\label{theorem:direct_obj_multiconvex}
\end{mytheorem}
Proof is given in Appendix \ref{appendix:theorem_direct_obj_multiconvex}.  As a
corollary, the objective function of FMs of arbitrary order is thus
multi-convex.  Theorem \ref{theorem:direct_obj_multiconvex} suggests that we can
minimize \eqref{eq:direct_obj} efficiently  when $\mathcal{K}=\mathcal{A}^m$ by
solving a succession of convex problems w.r.t.  $\bs{\lambda}$ and the rows of
$\bs{P}$.  We next show that when $m$ is odd, we can just fix $\bs{\lambda} =
\bs{1}$ without loss of generality.
\vspace{-0.1cm}
\begin{mylemma}{When is it useful to fit $\bs{\lambda}$?}

    Let $\mathcal{K} = \mathcal{H}^m \text{ or } \mathcal{A}^m$.  Then
\begin{align}
\min_{\bs{\lambda} \in \mathbb{R}^k, \bs{P} \in \mathbb{R}^{d \times k}}
D_{\mathcal{K}}(\bs{\lambda}, \bs{P}) &\le \min_{\bs{P} \in \mathbb{R}^{d
\times k}} D_{\mathcal{K}}(\bs{1}, \bs{P}) \quad \text{if } m \text{ is
even} \\
\min_{\bs{\lambda} \in \mathbb{R}^k, \bs{P} \in \mathbb{R}^{d \times k}}
D_{\mathcal{K}}(\bs{\lambda}, \bs{P}) &= \min_{\bs{P} \in \mathbb{R}^{d
\times k}} D_{\mathcal{K}}(\bs{1}, \bs{P}) \quad \text{if } m \text{ is
odd}.
\end{align}
\label{lemma:fit_lambda}
\end{mylemma}
\vspace{-1cm}
The result stems from the fact that $\mathcal{H}^m$ and $\mathcal{A}^m$ are
homogeneous functions.  If we define $\bs{v} \coloneqq \sign(\lambda)
\sqrt[m]{|\lambda|} \bs{p}$, then we obtain $\lambda \mathcal{H}^m(\bs{p},
\bs{x}) = \mathcal{H}^m(\bs{v}, \bs{x}) ~ \forall \lambda$ if $m$ is odd, and
similarly for $\mathcal{A}^m$. That is, $\lambda$ can be absorbed into $\bs{v}$
without loss of generality. When $m$ is even, $\lambda < 0$ cannot be absorbed
unless we allow complex numbers.  Because FMs fix $\bs{\lambda}=\bs{1}$, Lemma
\ref{lemma:fit_lambda} shows that the class of functions that FMs can represent
is possibly smaller than our framework.

\section{Lifted approach}
\label{sec:lifted_optimization}

\subsection{Conversion to low-rank tensor estimation problem}
\label{sec:lifted_conversion}

If we set $\mathcal{K} = \mathcal{H}^m$ in \eqref{eq:direct_obj}, the resulting
optimization problem is \textit{neither} convex \textit{nor} multi-convex w.r.t.
$\bs{P}$.  In \citep{convex_fm}, for $m=2$, it was proposed to cast parameter
estimation as a low-rank symmetric matrix estimation problem. A similar idea was
used in the context of phase retrieval in \citep{phase_lift}.  Inspired by these
works, we propose to convert the problem of estimating
$\bs{\lambda}$ and $\bs{P}$ to that of estimating a low-rank \textit{symmetric}
tensor $\tensor{W} \in \mathbb{S}^{d^m}$. 
Combined with a symmetrization trick,
this approach leads to an objective that is multi-convex, for \textit{both}
$\mathcal{K}=\mathcal{A}^m$ and $\mathcal{K}=\mathcal{H}^m$ (Section
\ref{sec:lifted_multi_convex}).

We begin by rewriting the kernel definitions using rank-one tensors.  For
$\homogkern{p}{x}{m}$, it is easy to see that 
\begin{equation}
    \homogkern{p}{x}{m} = 
    \langle \symtensor{\bs{p}}{m}, \symtensor{\bs{x}}{m} \rangle.
\label{eq:homogeneous_kernel_tensor}
\end{equation}
For $\combikern{p}{x}{m}$, we need to ignore irrelevant monomials. For
convenience, we introduce the following notation:
\begin{equation}
\langle \tensor{W}, \tensor{X} \rangle_> 
\coloneqq \sum_{j_m > \dots > j_1} \tensor{W}_{j_1, \dots, j_m} 
\tensor{X}_{j_1, \dots, j_m} \quad
\forall ~ \tensor{W}, \tensor{X} \in \mathbb{S}^{d^m}.
\end{equation}
We can now concisely rewrite the ANOVA kernel as
\begin{equation}
\combikern{p}{x}{m} =
\langle \symtensor{\bs{p}}{m}, \symtensor{\bs{x}}{m} \rangle_>.
\label{eq:combination_kernel_tensor}
\end{equation}

Our key insight is described
in the following lemma.
\begin{mylemma}{Link between tensors and kernel expansions}

Let $\tensor{W} \in \mathbb{S}^{d^m}$ have a symmetric outer product
decomposition
\citep{sym_tensors}
\vspace{-0.1cm}
\begin{equation}
\tensor{W} = \sum_{s=1}^k \lambda_s \symtensor{\bs{p}_s}{m}.
\label{eq:tensor_decomp}
\end{equation}
Let $\bs{\lambda} = [\lambda_1, \dots, \lambda_k]^\tr$ and
$\bs{P} = [\bs{p}_1, \dots, \bs{p}_k]$.  Then,
\begin{align}
\langle \tensor{W}, \symtensor{\bs{x}}{m} \rangle 
&= \hat{y}_{\mathcal{H}^m}(\bs{x}; \bs{\lambda}, \bs{P}) 
\quad \label{eq:homogkern_expansion} \\
\langle \tensor{W}, \symtensor{\bs{x}}{m} \rangle_>
&= \hat{y}_{\mathcal{A}^m}(\bs{x}; \bs{\lambda}, \bs{P}) 
.  \label{eq:combikern_expansion} 
\end{align}
\label{lemma:tensor_kernel}
\end{mylemma}
\vspace{-0.5cm}
The result follows immediately from \eqref{eq:homogeneous_kernel_tensor} and
\eqref{eq:combination_kernel_tensor}, and from the linearity of $\langle \cdot,
\cdot \rangle$ and $\langle \cdot, \cdot \rangle_>$. 
Given $\tensor{W} \in \mathbb{S}^{d^m}$, let us define the following objective
functions
\begin{align}
    L_{\mathcal{H}^m}(\tensor{W}) &\coloneqq \sum_{i=1}^n \ell\left(y_i, \langle
    \tensor{W}, \symtensor{\bs{x}_i}{m} \rangle \right) \\
    L_{\mathcal{A}^m}(\tensor{W}) &\coloneqq \sum_{i=1}^n \ell\left(y_i, \langle
    \tensor{W}, \symtensor{\bs{x}_i}{m} \rangle_> \right).
\end{align}
If $\tensor{W}$ is decomposed as in \eqref{eq:tensor_decomp}, then from Lemma
\ref{lemma:tensor_kernel}, we obtain $L_{\mathcal{K}}(\tensor{W}) =
D_{\mathcal{K}}(\bs{\lambda}, \bs{P})$ for $\mathcal{K}=\mathcal{H}^m$ or
$\mathcal{A}^m$.  This suggests that we can convert the problem of learning
$\bs{\lambda}$ and $\bs{P}$ to that of learning a symmetric tensor $\tensor{W}$
of (symmetric) rank $k$.  Thus, the problem of finding a small number of bases
$\bs{p}_1, \dots, \bs{p}_k$ and their associated weights $\lambda_1, \dots,
\lambda_k$ is converted to that of learning a \textit{low-rank} symmetric
tensor. Following \cite{phase_lift}, we call this approach \textit{lifted}.
Intuitively, we can think of $\tensor{W}$ as a tensor that contains the weights
for predicting $y$ of monomials of degree $m$. For instance, when $m=3$,
$\tensor{W}_{i,j,k}$ is the weight corresponding to the monomial $x_i x_j x_k$. 

\subsection{Multi-convex formulation}
\label{sec:lifted_multi_convex}

Estimating a low-rank symmetric tensor $\tensor{W} \in \mathbb{S}^{d^m}$ for
arbitrary integer $m \ge 2$ is in itself a difficult non-convex problem.
Nevertheless, based on a symmetrization trick, we can convert the problem to a
multi-convex one, which we can easily minimize by alternating minimization.
We first present our approach for the case $m=2$ to give intuitions then explain
how to extend it to $m \ge 3$.  

{\bf Intuition with the second-order case.} For the case $m=2$, we need to
estimate a low-rank symmetric matrix $\bs{W} \in \mathbb{S}^{d^2}$.  Naively
parameterizing $\bs{W}=\eigen{P}{\lambda}$ and solving for $\bs{\lambda}$ and
$\bs{P}$ does not lead to a multi-convex formulation for the case
$\mathcal{K}=\mathcal{H}^2$.  This is due to the fact that $\langle
\eigen{P}{\lambda}, \symtensor{\bs{x}}{2} \rangle$ is \textit{quadratic} in
$\bs{P}$. Our key idea is to parametrize $\bs{W} = \symop{\decomp{U}{V}}$ where
$\bs{U}, \bs{V} \in \mathbb{R}^{d \times r}$ and $\symop{\bs{M}} \coloneqq
\symmat{M} \in \mathbb{S}^{d^2}$ is the \textit{symmetrization} of $\bs{M} \in
\mathbb{R}^{d^2}$.  We then minimize
$L_{\mathcal{K}}(\symop{\decomp{U}{V}})$ w.r.t.  $\bs{U},\bs{V}$.

The main advantage is that both $\langle \symop{\decomp{U}{V}}, \cdot \rangle$
and $\langle \symop{\decomp{U}{V}}, \cdot \rangle_>$ are \textit{bi-linear} in
$\bs{U}$ and $\bs{V}$. This implies that
$L_{\mathcal{K}}(\symop{\decomp{U}{V}})$ is \textit{bi-convex} in $\bs{U}$ and
$\bs{V}$ and can therefore be efficiently minimized by alternating minimization.
Once we obtained $\bs{W} = \symop{\decomp{U}{V}}$, we can \textit{optionally}
compute its eigendecomposition $\bs{W} = \eigen{P}{\lambda}$, with
$k=\rank(\bs{W})$ and $r \le k \le 2r$, then apply
\eqref{eq:homogkern_expansion} or \eqref{eq:combikern_expansion} to obtain the
model in kernel expansion form.

{\bf Extension to higher-order case.} For $m \ge 3$, we now estimate a low-rank
symmetric tensor $\tensor{W} = \symop{\tensor{M}} \in \mathbb{S}^{d^m}$, where
$\tensor{M} \in \mathbb{R}^{d^m}$ and $\symop{\tensor{M}}$ is the symmetrization
of $\tensor{M}$ (cf. Appendix \ref{appendix:proof_sym_forms}). We decompose
$\tensor{M}$ using $m$ matrices of size $d \times r$.  Let us call these
matrices $\{\bs{U}^t\}_{t=1}^m$ and their columns $\bs{u}^t_s = [u^t_{1s},
\dots, u^t_{ds}]^\tr$. Then the decomposition of $\tensor{M}$ can be expressed
as a sum of rank-one tensors
\vspace{-0.1cm}
\begin{equation}
    \tensor{M} = \sum_{s=1}^r \bs{u}^1_s \otimes \dots \otimes \bs{u}^m_s.
\label{eq:decomp_M}
\end{equation}
Due to multi-linearity of \eqref{eq:decomp_M} w.r.t.  $\bs{U}^1,
\dots, \bs{U}^m$, the objective function $L_\mathcal{K}$ is
\textit{multi-convex} in $\bs{U}^1, \dots, \bs{U}^m$.  

{\bf Computing predictions efficiently.} When $\mathcal{K}=\mathcal{H}^m$,
predictions are computed by $\langle \tensor{W}, \symtensor{\bs{x}}{m} \rangle$.
To compute them efficiently, we use the following lemma.
\begin{mylemma}{Symmetrization does not affect inner product}
\begin{equation}
    \langle \symop{\tensor{M}}, \tensor{X} 
    \rangle = \langle \tensor{M}, \tensor{X} \rangle \quad
\forall ~ \tensor{M} \in \mathbb{R}^{d^m}, \tensor{X} \in \mathbb{S}^{d^m}, m
\ge 2.
\label{eq:sym_forms}
\end{equation}
\label{lemma:sym_forms}
\end{mylemma}
\vspace{-0.5cm}
Proof is given in Appendix \ref{appendix:proof_sym_forms}.  
Using
$\symtensor{\bs{x}}{m} \in \mathbb{S}^{d^m}$, $\tensor{W} =
\symop{\tensor{M}}$ and \eqref{eq:decomp_M}, we then obtain
\begin{equation}
\langle \tensor{W}, \symtensor{\bs{x}}{m} \rangle = 
\langle \tensor{M}, \symtensor{\bs{x}}{m} \rangle 
= \sum_{s=1}^r \prod_{t=1}^m \langle \bs{u}_s^t, \bs{x} \rangle.
\end{equation}
As a result, we never need to explicitly compute the symmetrized tensor.  For
the case
$\mathcal{K}=\mathcal{A}^2$, cf.  Appendix \ref{appendix:lifted_cd_anova}.

\section{Regularization}
\label{sec:regularization}

In some applications, the number of bases or the rank constraint are not enough
for obtaining good generalization performance and it is necessary to consider
additional form of regularization.  For the lifted objective with
$\mathcal{K}=\mathcal{H}^2$ or $\mathcal{A}^2$, we use the typical
Frobenius-norm regularization
\vspace{-0.1cm}
\begin{equation}
    \tilde{L}_{\mathcal{K}}(\bs{U}, \bs{V}) \coloneqq 
    L_{\mathcal{K}}(\symop{\decomp{U}{V}}) + 
\frac{\beta}{2} \sumfrobnorm{U}{V},
\label{eq:lifted_regul_obj}
\end{equation}
where $\beta > 0$ is a regularization hyper-parameter.  
For the direct objective, we introduce the new regularization
\vspace{-0.3cm}
\begin{equation}
    \tilde{D}_{\mathcal{K}}(\bs{\lambda}, \bs{P}) \coloneqq 
D_{\mathcal{K}}(\bs{\lambda}, \bs{P}) + 
\beta \sum_{s=1}^k |\lambda_s| ~ \|\bs{p}_s\|^2.
\label{eq:direct_regul_obj}
\end{equation}
This allows us to regularize $\bs{\lambda}$ and $\bs{P}$ with a single
hyper-parameter. Let us define the following nuclear norm penalized objective:
\begin{equation}
    \bar{L}_{\mathcal{K}}(\bs{M}) \coloneqq 
    L_{\mathcal{K}}(\symop{\bs{M}}) + \beta \|\bs{M}\|_*.
\label{eq:nuclear_obj}
\end{equation}
We can show that \eqref{eq:lifted_regul_obj}, \eqref{eq:direct_regul_obj} and
\eqref{eq:nuclear_obj} are equivalent in the following sense.
\vspace{-0.2cm}
\begin{mytheorem}{Equivalence of regularized problems}

    Let $\mathcal{K}=\mathcal{H}^2$ or $\mathcal{A}^2$, then
\begin{equation}
\min_{\bs{M} \in \mathbb{R}^{d^2}}
\bar{L}_{\mathcal{K}}(\bs{M}) =
\min_{\substack{\bs{U} \in \mathbb{R}^{d \times r}\\\bs{V} \in \mathbb{R}^{d \times r}}}
\tilde{L}_{\mathcal{K}}(\bs{U}, \bs{V}) 
= \min_{\substack{\bs{\lambda} \in \mathbb{R}^k\\\bs{P} \in \mathbb{R}^{d
\times k}}}
\tilde{D}_{\mathcal{K}}(\bs{\lambda}, \bs{P})
\end{equation}
where $\rank(\bs{M}^*) \le r=k$ and $\bs{M}^* \in \displaystyle{\argmin_{\bs{M}
\in \mathbb{R}^{d^2}}}
~ \bar{L}_{\mathcal{K}}(\bs{M})$.
\label{theorem:equivalence_reg_problems}
\end{mytheorem}
Proof is given in Appendix \ref{appendix:proof_equivalence_reg_problems}.  Our
proof relies on the variational form of the nuclear norm and is thus limited to
$m=2$.  One of the key ingredients of the proof is to show that the
minimizer of \eqref{eq:nuclear_obj} is always a symmetric matrix. 
In addition to Theorem \ref{theorem:equivalence_reg_problems}, from
\citep{abernethy}, we also know that every local minimum $\bs{U},\bs{V}$ of
\eqref{eq:lifted_regul_obj} gives a global solution $\decomp{U}{V}$ of
\eqref{eq:nuclear_obj} provided that $\rank(\bs{M}^*) \le r$.  Proving a similar
result for \eqref{eq:direct_regul_obj} is a future work.  When $m \ge 3$, as
used in our experiments, a squared Frobenius norm penalty on $\bs{P}$ (direct
objective) or on $\{\bs{U}^t\}_{t=1}^m$ (lifted objective) works well in
practice, although we lose the theoretical connection with the nuclear norm. 

\section{Coordinate descent algorithms}
\label{sec:coordinate_descent}

We now describe how to learn the model parameters by coordinate descent, which
is a state-of-the-art \textit{learning-rate free} solver for multi-convex
problems (e.g., \citet{mc_cd}). In the following, we assume that $\ell$ is
$\mu$-smooth.

{\bf Direct objective with $\mathcal{K}=\mathcal{A}^m$ for $m \in \{2,3\}$.}
First, we note that minimizing \eqref{eq:direct_regul_obj} w.r.t. $\bs{\lambda}$
can be reduced to a standard $\ell_1$-regularized convex objective via a simple
change of variable.  Hence we focus on minimization w.r.t. $\bs{P}$.  

Let us denote the
elements of $\bs{P}$ by $p_{js}$.  Then, our algorithm cyclically performs the
following update for
all $s \in [k]$ and $j \in [d]$:
\begin{equation}
p_{js} \leftarrow p_{js} - \eta^{-1} \left[ \sum_{i=1}^n \ell'(y_i,
    \hat{y}_i) \partialfrac{\hat{y}_i}{p_{js}} + 2 \beta |\lambda_s| p_{js}
    \right],
\label{eq:update_direct}
\end{equation}
where $\eta \coloneqq \mu \sum_{i=1}^n
\left(\partialfrac{\hat{y}_i}{p_{js}}\right)^2 + 2 \beta |\lambda_s|$.  Note
that when $\ell$ is the squared loss, the above is equivalent to a Newton update
and is the exact coordinate-wise minimizer. 

The key challenge to use CD is
computing
$\partialfrac{\hat{y}_i}{p_{js}} = \lambda_s
\partialfrac{\mathcal{A}^m(\bs{p}_s, \bs{x}_i)}{p_{js}}$ efficiently.  Let us
denote the elements of $\bs{X}$ by $x_{ji}$. 
Using Lemma \ref{lemma:linear_time_anova}, we obtain
$\partialfrac{\mathcal{A}^2(\bs{p}_s, \bs{x}_i)}{p_{js}} = \langle \bs{p}_s,
\bs{x}_i \rangle x_{ji} - p_{js} x_{ji}^2$ and
$\partialfrac{\mathcal{A}^3(\bs{p}_s, \bs{x}_i)}{p_{js}} =
\mathcal{A}^2(\bs{p}_s, \bs{x}_i) x_{ji} - p_{js} x_{ji}^2 \langle \bs{p}_s,
\bs{x}_i \rangle + p_{js}^2 x_{ji}^3$. If for all $i \in [n]$ and for $s$ fixed,
we maintain $\langle \bs{p}_s, \bs{x}_i \rangle$ and $\mathcal{A}^2(\bs{p}_s,
\bs{x}_i)$ (i.e., keep in sync after every update of $p_{js}$), then computing
$\partialfrac{\hat{y}_i}{p_{js}}$ takes $O(m)$ time. Hence the cost of one
epoch, i.e.  updating all elements of $\bs{P}$ once, is $O(mk n_z(\bs{X}))$.
Complete details and pseudo code are given in Appendix \ref{appendix:direct_cd}.  

To our knowledge, this is the first CD algorithm capable of training third-order
FMs.  Supporting arbitrary $m \in \mathbb{N}$ is an important future work.

{\bf Lifted objective with $\mathcal{K}=\mathcal{H}^m$.} Recall that we want to
learn the matrices $\{\bs{U}^t\}_{t=1}^m$, whose columns we denote by
$\bs{u}^t_s = [u^t_{1s}, \dots, u^t_{ds}]^\tr$. Our algorithm cyclically
performs the following update for all $t \in [m]$, $s \in [r]$ and $j \in [d]$:
\begin{equation}
u_{js}^t \leftarrow u_{js}^t - \eta^{-1} \left[\sum_{i=1}^n
    \ell'(y_i, \hat{y}_i) \partialfrac{\hat{y}_i}{u_{js}^t} + \beta
    u^t_{js}\right],
\label{eq:update_lifted}
\end{equation}
where $\eta \coloneqq \mu \sum_{i=1}^n
\left(\partialfrac{\hat{y}_i}{u_{js}^t}\right)^2 + \beta$.  
The main difficulty
is computing $\partialfrac{\hat{y}_i}{u_{js}^t} = \prod_{t' \neq t} \langle
\bs{u}^{t'}_s, \bs{x}_i \rangle x_{ji}$ efficiently. If for all $i \in [n]$ and
for $t$ and $s$ fixed, we maintain $\xi_i \coloneqq \prod_{t' \neq t} \langle
\bs{u}^{t'}_s, \bs{x}_i \rangle$, then the cost of computing
$\partialfrac{\hat{y}_i}{u_{js}^t}$ is $O(1)$.  Hence the cost of one epoch is
$O(mr n_z(\bs{X}))$, the same as SGD.  Complete details are given
in Appendix \ref{appendix:lifted_cd}.  

{\bf Convergence.} The above updates decrease the objective
\textit{monotonically}.  Convergence to a stationary point is guaranteed
following \citep[Proposition 2.7.1]{bertsekas}.

\vspace{-0.2cm}
\section{Inhomogeneous polynomial models}
\label{sec:inhomogeneous}

The algorithms presented so far are designed for homogeneous polynomial kernels
$\mathcal{H}^m$ and $\mathcal{A}^m$. These kernels only use monomials of the
\textit{same} degree $m$. However, in many applications, we would like to use
monomials of \textit{up to} some degree.  In this section, we propose a simple
idea to do so using the algorithms presented so far, unmodified. Our key
observation is that we can easily turn homogeneous polynomials into
inhomogeneous ones by augmenting the dimensions of the training data with
dummy features. 

We begin by explaining how to learn inhomogeneous polynomial models using
$\mathcal{H}^m$.  Let us denote $\bs{\tilde{p}}^\tr \coloneqq [\gamma,
\bs{p}^\tr] \in \mathbb{R}^{d+1}$ and $\bs{\tilde{x}}^\tr \coloneqq [1,
\bs{x}^\tr] \in \mathbb{R}^{d+1}$.  Then, we obtain
\begin{equation}
    \mathcal{H}^m(\bs{\tilde{p}}, \bs{\tilde{x}}) = 
\langle \bs{\tilde{p}}, \bs{\tilde{x}} \rangle^m =
    (\gamma + \innerprod{p}{x})^m =
    \polykern{p}{x}{m}{\gamma}.
\end{equation}
Therefore, if we prepare the augmented training set $\bs{\tilde{x}}_1, \dots,
\bs{\tilde{x}}_n$, the problem of learning a model of the form $\sum_{s=1}^k
\lambda_s \mathcal{P}^m_{\gamma_s}(\bs{p}_s,\bs{x})$ can be converted to that of
learning a rank-$k$ symmetric tensor $\tensor{W} \in \mathbb{S}^{(d+1)^m}$ using
the method presented in Section \ref{sec:lifted_optimization}. Note that the
parameter $\gamma_s$ is automatically learned from data for each basis
$\bs{p}_s$.

Next, we explain how to learn inhomogeneous polynomial models using
$\mathcal{A}^m$.  Using Lemma \ref{lemma:multi_linearity}, we immediately
obtain for $1 \le m \le d$:
\begin{equation}
\combikern{\tilde{p}}{\tilde{x}}{m}= \combikern{p}{x}{m} + \gamma
\combikern{p}{x}{m-1}.
\label{eq:combikern_augment}
\end{equation}
For instance, when $m=2$, we obtain
\begin{equation}
\combikern{\tilde{p}}{\tilde{x}}{2} 
= \combikern{p}{x}{2} + \gamma \combikern{p}{x}{1} 
= \combikern{p}{x}{2} + \gamma \innerprod{p}{x}.
\end{equation}
Therefore, if we prepare the augmented training set $\bs{\tilde{x}}_1, \dots,
\bs{\tilde{x}}_n$, we can easily learn a combination of linear kernel and
second-order ANOVA kernel using methods presented in Section
\ref{sec:direct_optimization} or Section \ref{sec:lifted_optimization}.  
Note that \eqref{eq:combikern_augment} only states the relation between two
ANOVA kernels of consecutive degrees. Fortunately, we can also apply
\eqref{eq:combikern_augment} recursively. Namely, by adding $m-1$ dummy
features, we can sum the kernels from $\mathcal{A}^m$ down to $\mathcal{A}^1$
(i.e., linear kernel).

\vspace{-0.2cm}
\section{Experimental results}
\label{sec:experiments}

In this section, we present experimental results, focusing on regression tasks.
Datasets are described in Appendix \ref{appendix:datasets}.
In all experiments, we set $\ell(y, \hat{y})$ to the squared loss.

\subsection{Direct optimization: is it useful to fit $\bs{\lambda}$?}
\label{sec:exp_fit_lambda}

As explained in Section \ref{lemma:fit_lambda}, there is no benefit to fitting
$\bs{\lambda}$ when $m$ is odd, since $\mathcal{A}^m$ and $\mathcal{H}^m$ can
absorb $\bs{\lambda}$ into $\bs{P}$. This is however not the case when $m$ is
even: $\mathcal{A}^m$ and $\mathcal{H}^m$ can absorb absolute values but not
negative signs (unless complex numbers are allowed for parameters). Therefore,
when $m$ is even, the class of functions we can represent with models of the
form \eqref{eq:prediction_function} is possibly smaller if we fix
$\bs{\lambda}=\bs{1}$ (as done in FMs). 

\begin{figure}[t]
    \center
    \subfigure[$\mathcal{K}=\mathcal{A}^2$]{
    \includegraphics[scale=0.18]{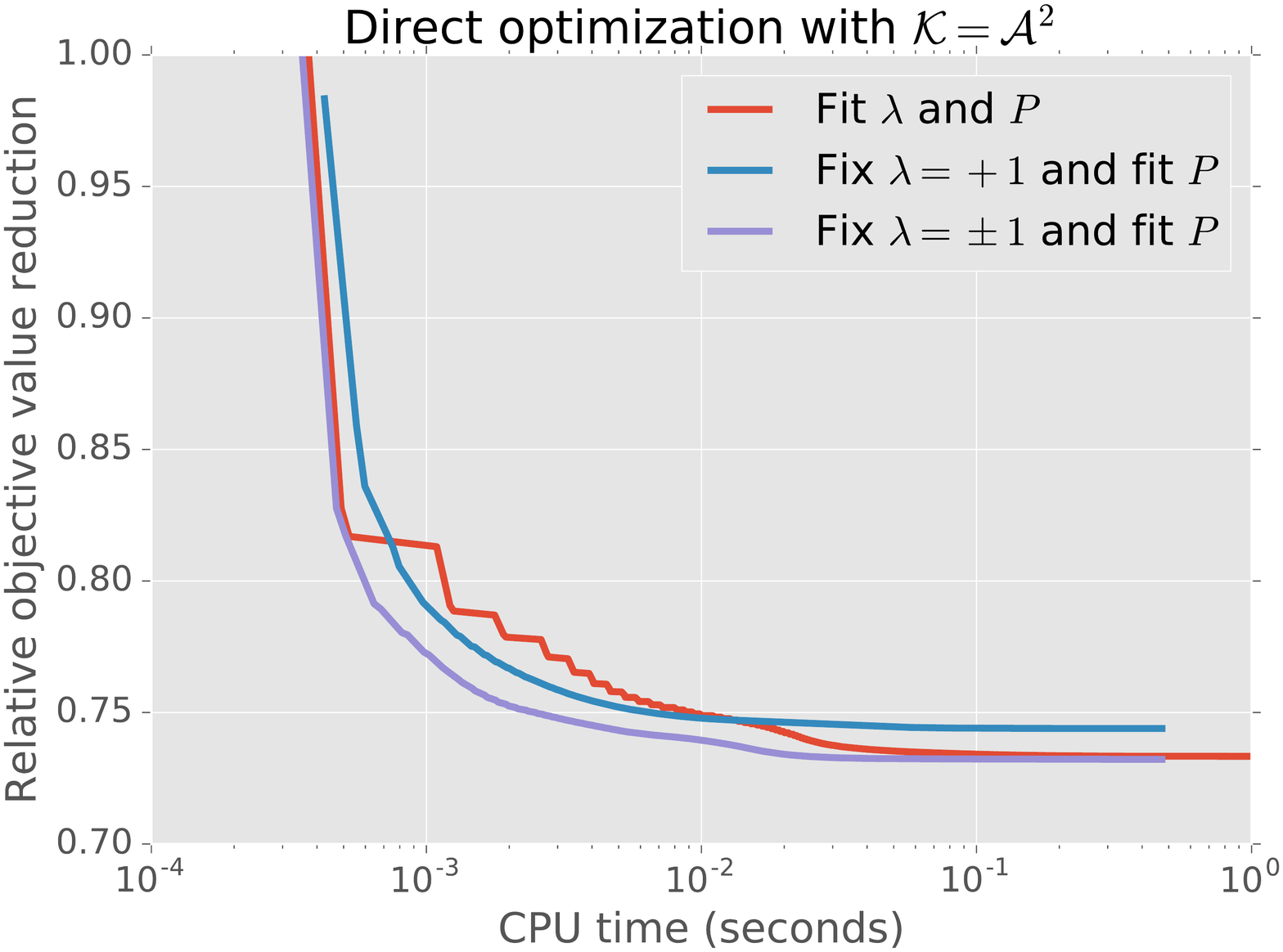}
}
    \subfigure[$\mathcal{K}=\mathcal{H}^2$]{
    \includegraphics[scale=0.18]{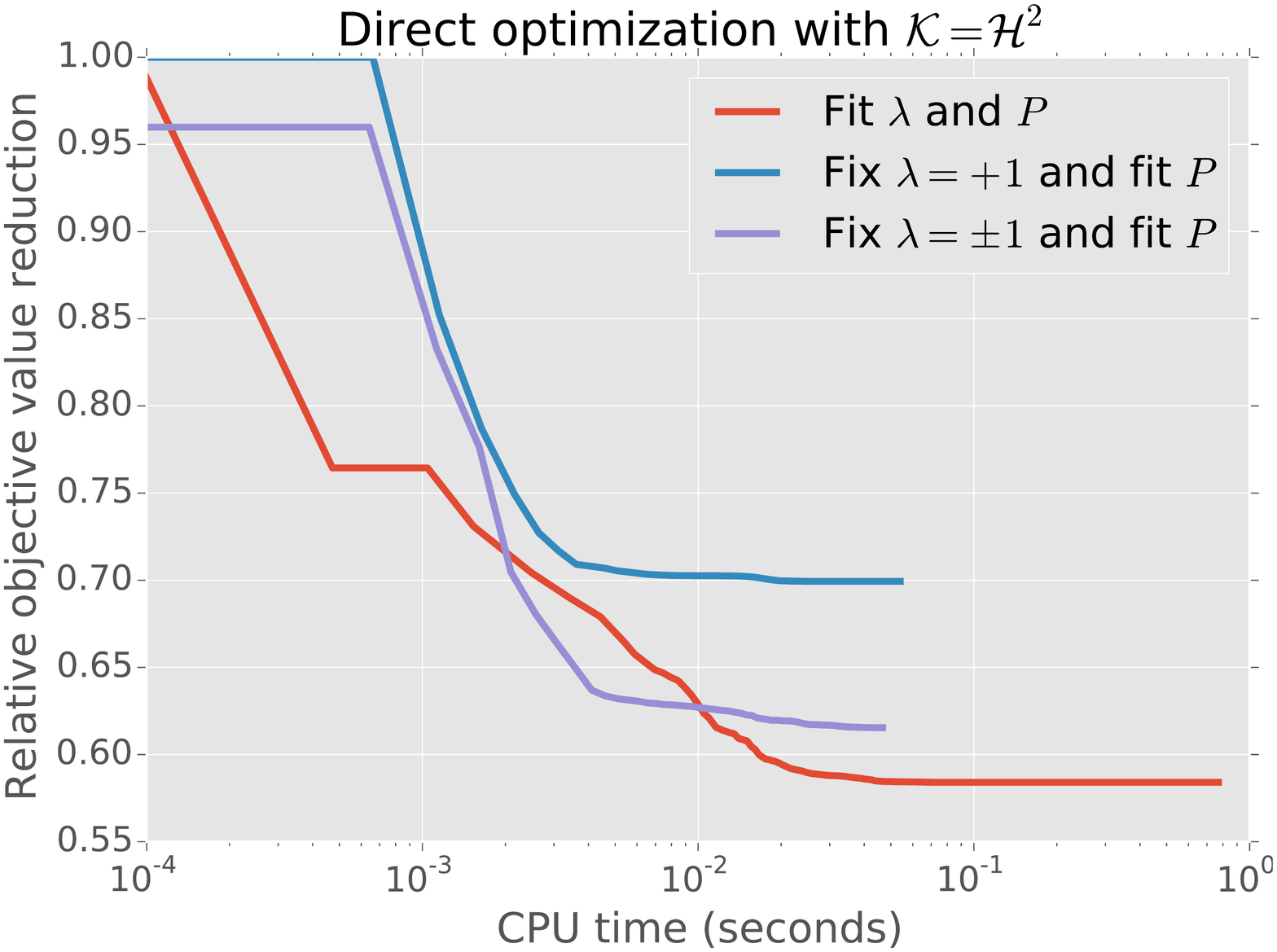}
    }
    \caption{{\bf Effect of fitting $\bs{\lambda}$ when using direct
        optimization} on the \textit{diabetes} dataset with $m=2$, $\beta=10$
        and $k=4$.  Objective values were computed by \eqref{eq:direct_regul_obj}
        and were normalized by the worst initialization's objective value.}
\label{figure:direct_cmp}
\end{figure}

To check that this is indeed the case, 
on the \textit{diabetes} dataset, we minimized 
\eqref{eq:direct_regul_obj} with $m=2$ as follows:
\begin{itemize}[topsep=0pt,itemsep=-1ex,partopsep=1ex,parsep=1ex]
    \item[a)] minimize w.r.t. both $\bs{\lambda}$ and $\bs{P}$ alternatingly,
    \item[b)] fix $\lambda_s = 1$ for $s \in [k]$ and minimize w.r.t. $\bs{P}$,
    \item[c)] fix $\lambda_s = \pm 1$ with proba. $0.5$ and minimize w.r.t. $\bs{P}$.
    \end{itemize}
We initialized elements of $\bs{P}$ by $p_{js} \sim \mathcal{N}(0, 0.01)$ for
all $j \in [d]$, $s \in [k]$.  Our results are shown in Figure
\ref{figure:direct_cmp}.  
For $\mathcal{K}=\mathcal{A}^2$, we use CD and for $\mathcal{K}=\mathcal{H}^2$,
we use L-BFGS.
Note that since \eqref{eq:direct_regul_obj} is convex w.r.t.
$\bs{\lambda}$, a)  is insensitive to the initialization of $\bs{\lambda}$ as
long as we fit $\bs{\lambda}$ before $\bs{P}$.  Not surprisingly, fitting
$\bs{\lambda}$ allows us to achieve a smaller objective value. This is
especially apparent when $\mathcal{K}=\mathcal{H}^2$. However, the difference is
much smaller when $\mathcal{K}=\mathcal{A}^2$.  We give intuitions as to why
this is the case in Section \ref{sec:discussion}.

We emphasize that this experiment was designed to confirm that fitting
$\bs{\lambda}$ does indeed improve representation power of the model when $m$ is
even. In practice, it is possible that fixing $\bs{\lambda}=\bs{1}$ reduces
overfitting and thus improves \textit{generalization error}. However, this
highly depends on the data.

\subsection{Direct vs. lifted optimization}
\label{sec:exp_direct_vs_lifted}

\begin{figure}[t]
    \center
    \subfigure[$\mathcal{K}=\mathcal{A}^2$]{
    \includegraphics[scale=0.18]{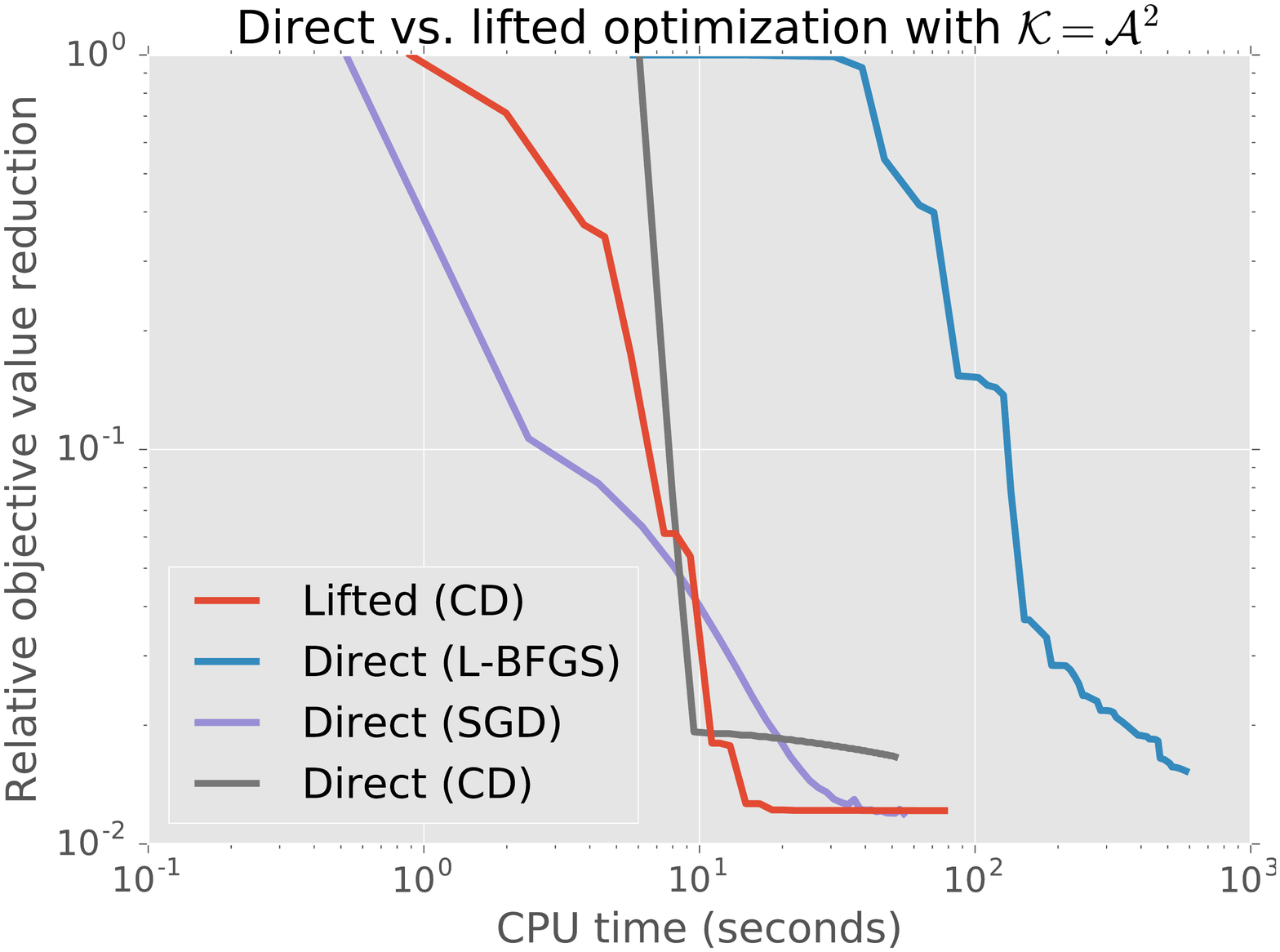}
}
    \subfigure[$\mathcal{K}=\mathcal{H}^2$]{
    \includegraphics[scale=0.18]{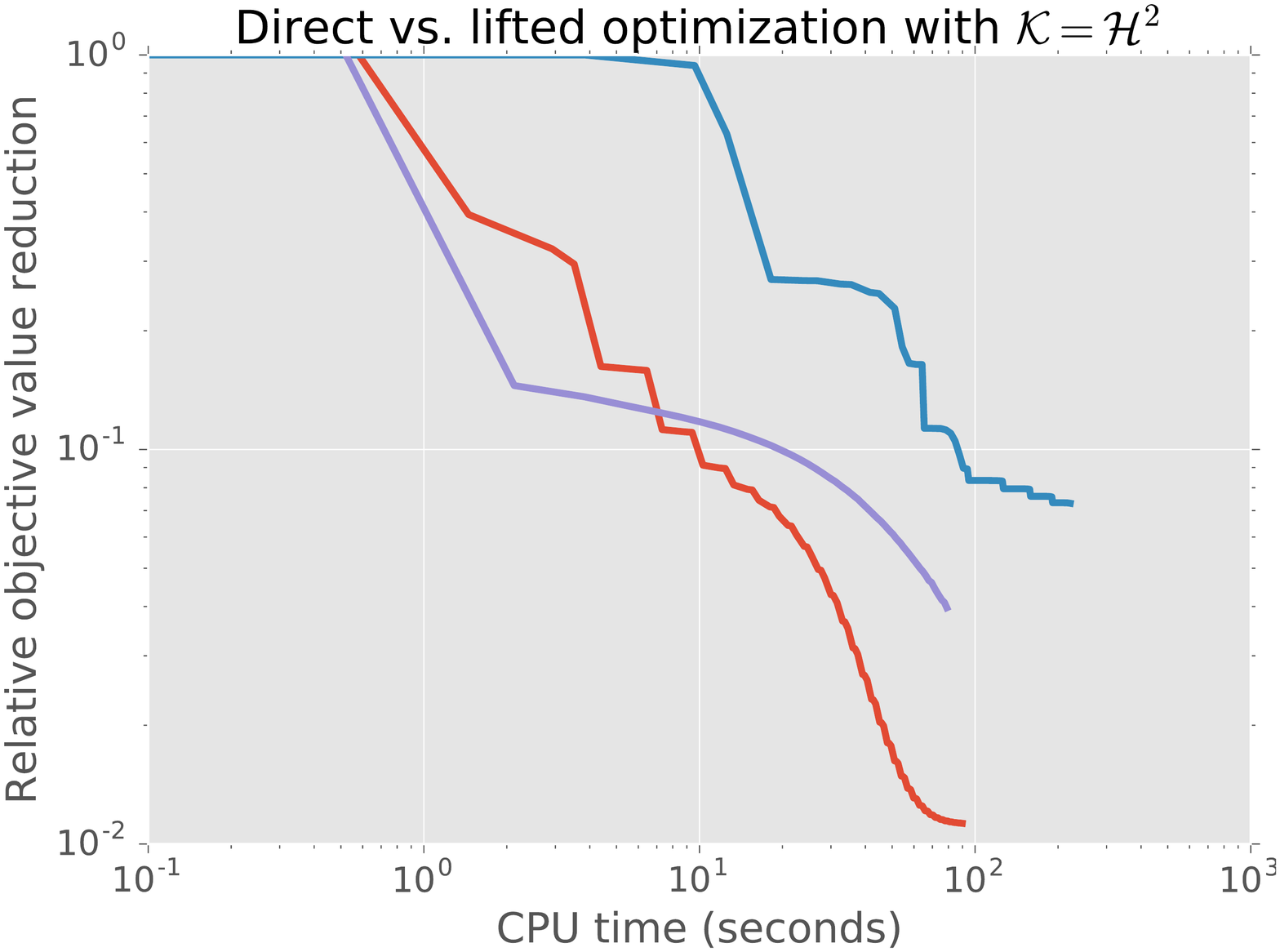}
}
\caption{{\bf Comparison of the direct and lifted optimization approaches} on the
    \textit{E2006-tfidf} high-dimensional dataset with $m=2$, $\beta=100$ and $r=k/2=10$.
In order to learn an inhomogeneous polynomial, we added a dummy feature to all
training instances, as explained in Section \ref{sec:inhomogeneous}.  Objective
values shown were computed by \eqref{eq:direct_regul_obj} and
\eqref{eq:lifted_regul_obj} and normalized by the initialization's objective
value.
}
\label{figure:direct_vs_lifted}
\end{figure}

In this section, we compare the direct and lifted optimization approaches on
high-dimensional data when $m=2$. To compare the two approaches fairly, we
propose the following initialization scheme.  Recall that, at the end of the
day, both approaches are essentially learning a low rank symmetric matrix:
$\bs{W} = \symop{\decomp{U}{V}}$ for lifted and $\bs{W} = \eigen{P}{\lambda}$
for direct optimization. This suggests that we can easily convert the matrices
$\bs{U},\bs{V} \in \mathbb{R}^{d \times r}$ used for initializing lifted
optimization to $\bs{P} \in \mathbb{R}^{d \times k}$ and $\bs{\lambda} \in
\mathbb{R}^{d \times k}$ by computing the (reduced) eigendecomposition of
$\symop{\decomp{U}{V}}$. Note that because we solve the lifted optimization
problem by coordinate descent, $\decomp{U}{V}$ is never symmetric and therefore
the rank of $\symop{\decomp{U}{V}}$ is usually twice that of $\decomp{U}{V}$.
Hence, in practice, we have that $r=k/2$.  In our experiment, we compared four
methods: lifted objective solved by CD, direct objective solved by CD, L-BFGS
and SGD.  For lifted optimization, we initialized the elements of $\bs{U}$ and
$\bs{V}$ by sampling from $\mathcal{N}(0, 0.01)$. For direct optimization, we
obtained $\bs{P}$ and $\bs{\lambda}$ as explained.  Results on the
\textit{E2006-tfidf} high-dimensional dataset are shown in Figure
\ref{figure:direct_vs_lifted}.  For $\mathcal{K}=\mathcal{A}^2$, we find that
Lifted (CD) and Direct (CD) have similar convergence speed and both outperform
Direct (L-BFGS). For $\mathcal{K}=\mathcal{H}^2$, we find that Lifted (CD)
outperforms both Direct (L-BFGS) and Direct (SGD). Note that we did not
implement Direct (CD) for $\mathcal{K}=\mathcal{H}^2$ since the direct
optimization problem is not coordinate-wise convex, as explained in Section
\ref{sec:lifted_optimization}.

\vspace{-0.3cm}
\subsection{Recommender system experiment}
\label{sec:exp_recsys}

\begin{figure}[t]
    \center
\subfigure[Last.fm]{
    \includegraphics[scale=0.18]{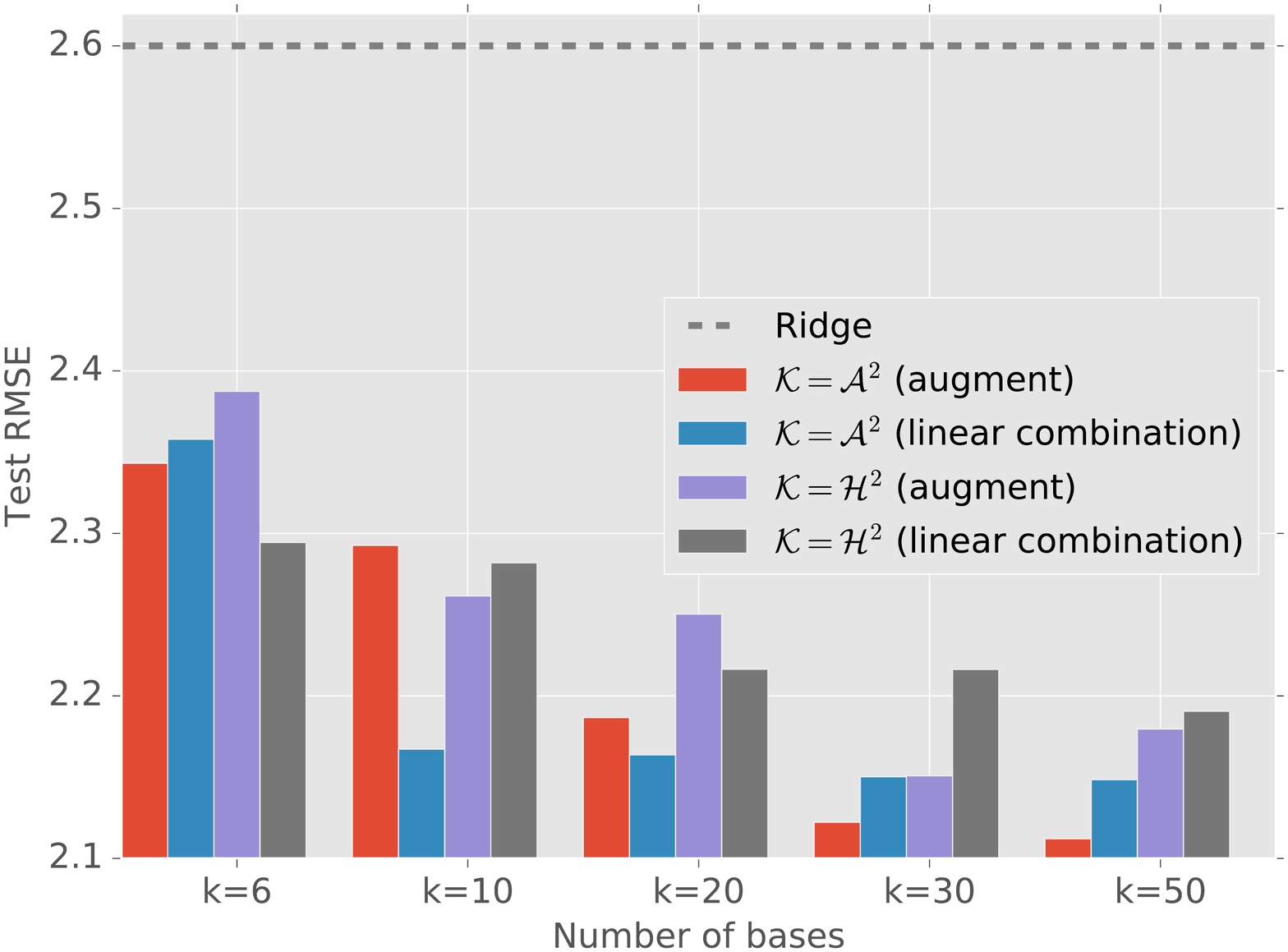}
}
\subfigure[Movielens 1M]{
    \includegraphics[scale=0.18]{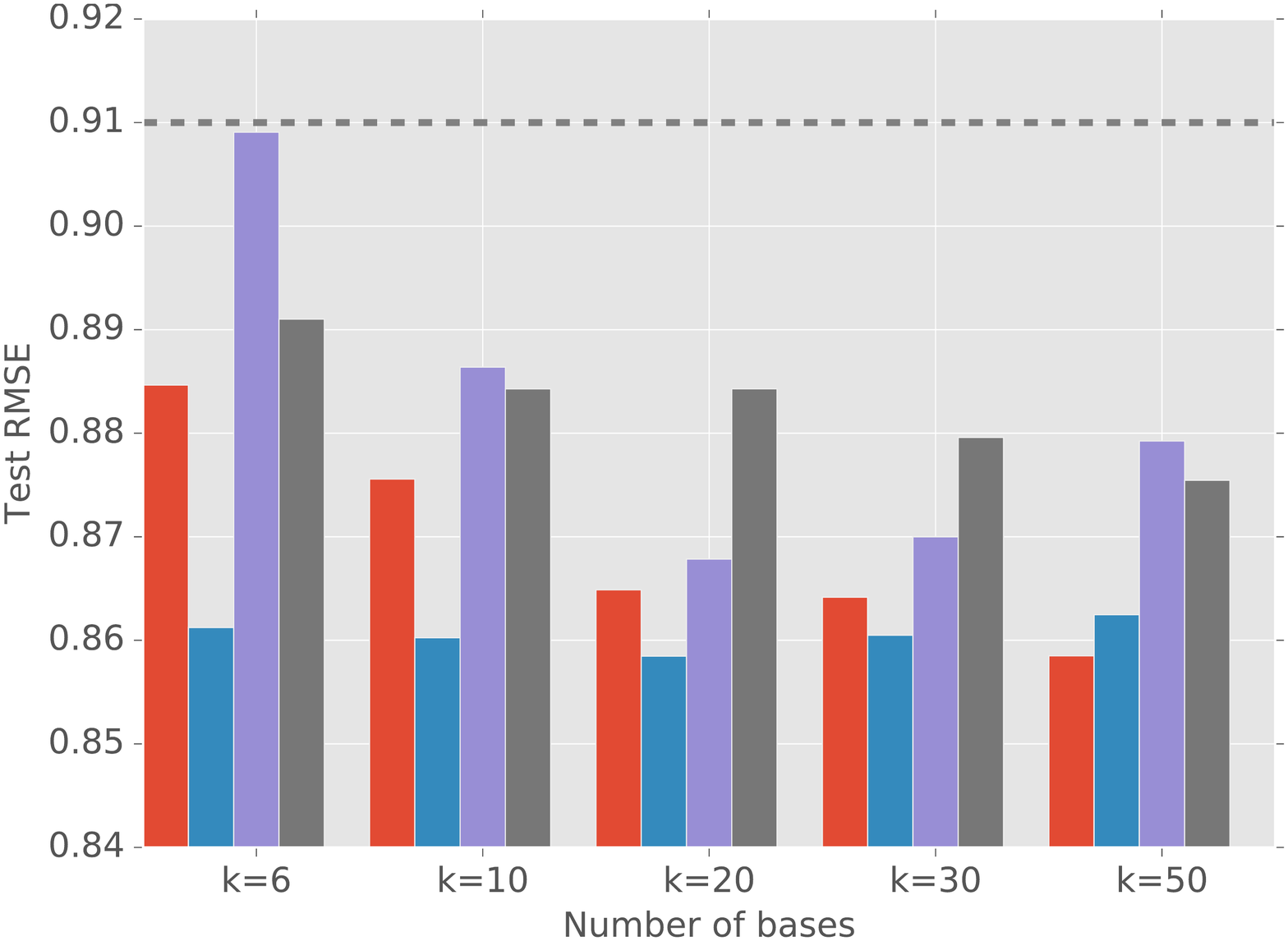}
}
\caption{{\bf Predicted rating error} on the \textit{Last.fm} and
\textit{Movielens 1M} datasets. The metric used is RMSE on the test set (lower
is better). The hyper-parameter $\beta$ was selected from 10 log-spaced values
in the interval $[10^{-3}, 10^3]$ by $5$-fold cross-validation.}
\label{figure:recsys}
\end{figure}

To confirm the ability of the proposed framework to infer the weights of
unobserved feature interactions, we conducted experiments on \textit{Last.fm}
and \textit{Movielens 1M}, two standard recommender system datasets. Following
\citep{libfm}, matrix factorization can be reduced to FMs by creating a
dataset of $(\bs{x}_i, y_i)$ pairs where $\bs{x}_i$ contains the one-hot
encoding of the user and item and $y_i$ is the corresponding rating (i.e.,
number of training instances equals number of ratings).  We compared four
models:
\begin{itemize}[topsep=0pt,itemsep=-1ex,partopsep=1ex,parsep=1ex]
    \item[a)] $\mathcal{K}=\mathcal{A}^2$ (augment): $\hat{y} =
        \hat{y}_{\mathcal{A}^2}(\bs{\tilde{x}})$, with
$\bs{\tilde{x}}^\tr \coloneqq [1, \bs{x}^\tr]$,
    \item[b)] $\mathcal{K}=\mathcal{A}^2$ (linear combination): $\hat{y} =
        \langle \bs{w}, \bs{x} \rangle + \hat{y}_{\mathcal{A}^2}(\bs{x})$,
    \item[c)] $\mathcal{K}=\mathcal{H}^2$ (augment): $\hat{y} =
        \hat{y}_{\mathcal{H}^2}(\bs{\tilde{x}})$ and
    \item[d)] $\mathcal{K}=\mathcal{H}^2$ (linear combination): $\hat{y} =
        \langle \bs{w}, \bs{x} \rangle + \hat{y}_{\mathcal{H}^2}(\bs{x})$,
    \end{itemize}
where $\bs{w} \in \mathbb{R}^d$ is a vector of first-order weights, estimated
from training data. Note that b)
and d) are exactly the same as FMs and PNs, respectively.  
Results are shown in Figure \ref{figure:recsys}. We see that $\mathcal{A}^2$
tends to outperform $\mathcal{H}^2$ on these tasks.  We
hypothesize that this the case because features are binary (cf.,  discussion in
Section \ref{sec:discussion}). We also see that simply augmenting the features
as suggested in Section \ref{sec:inhomogeneous} is comparable or better than
learning additional first-order feature weights, as done in FMs and PNs.

\subsection{Low-budget non-linear regression experiment}
\label{sec:exp_low_budget}

\begin{figure*}[t]
    \center
    \subfigure[abalone]{
    \includegraphics[scale=0.17]{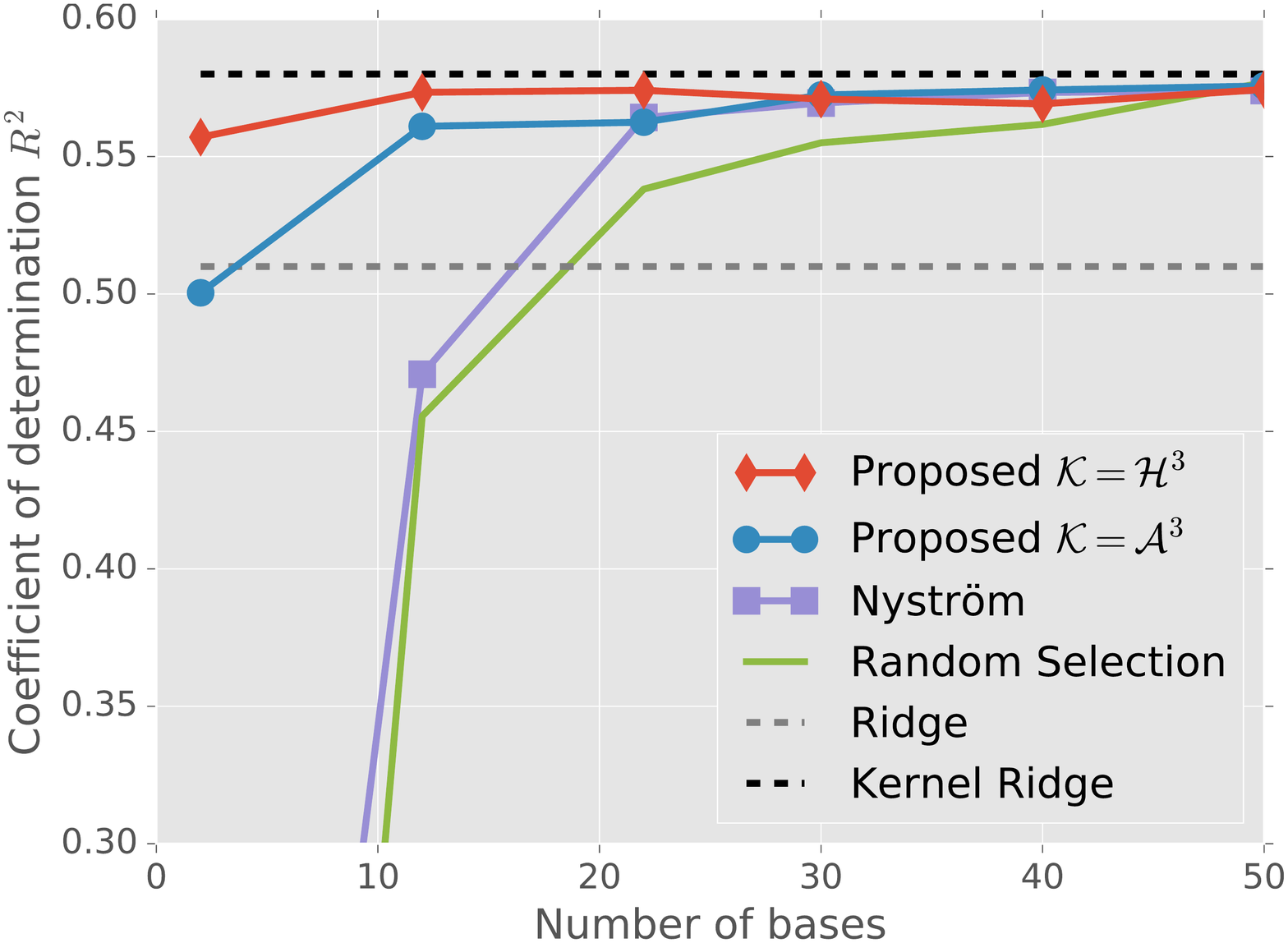}
}
    \subfigure[cadata]{
    \includegraphics[scale=0.17]{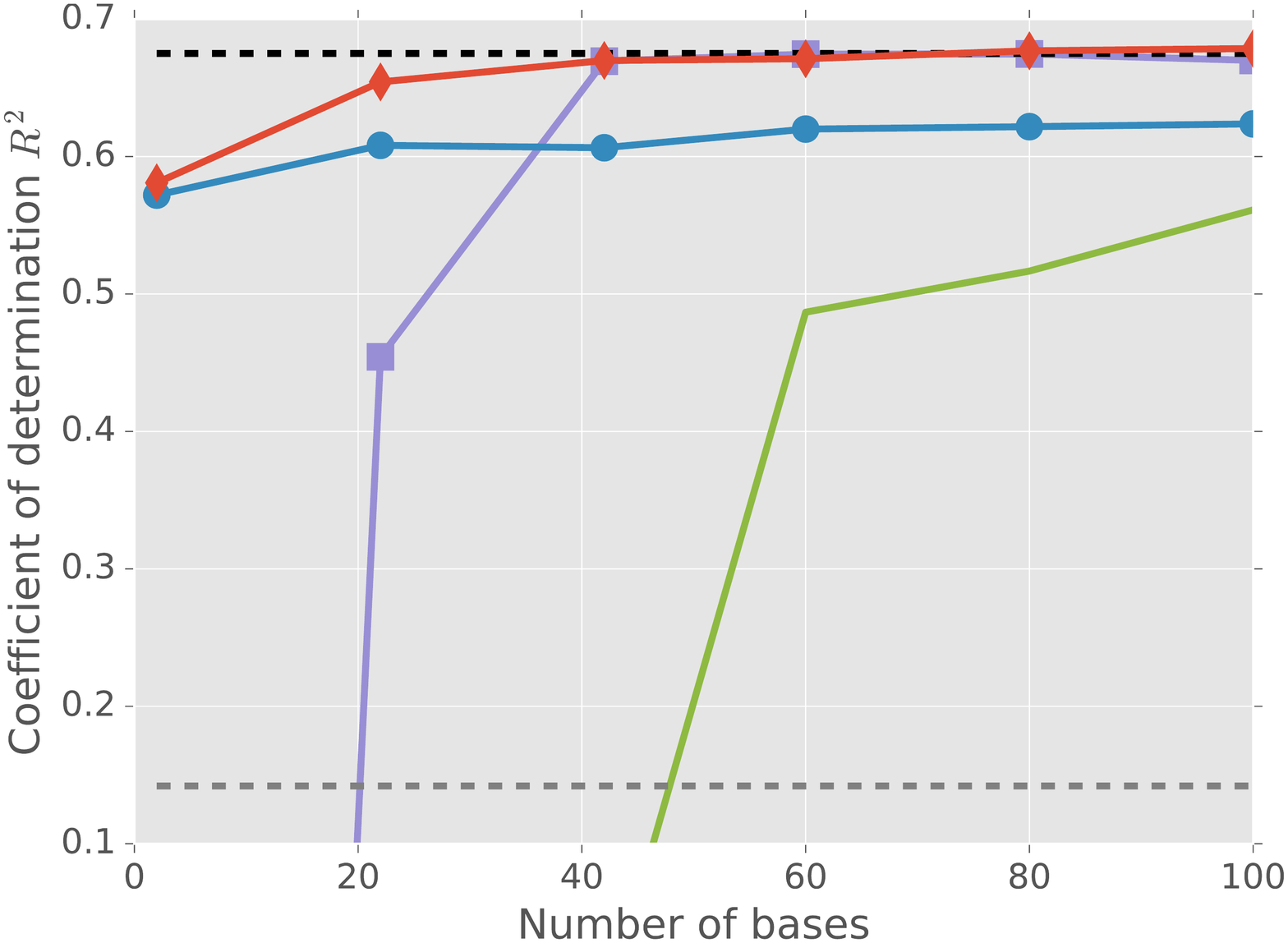}
}
    \subfigure[cpusmall]{
    \includegraphics[scale=0.17]{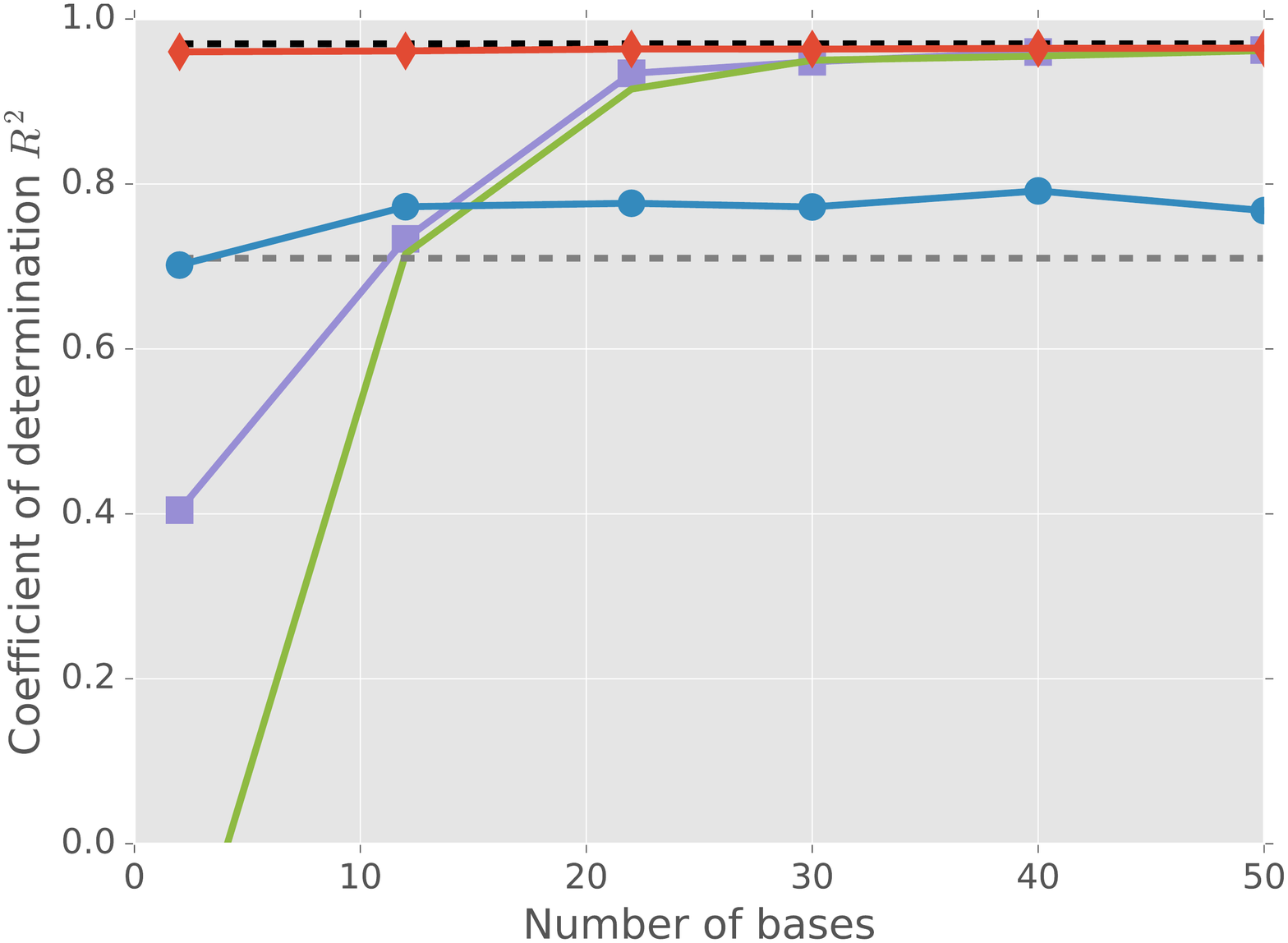}
}
\caption{{\bf Regression performance as a function of the number of bases.} The
    metric used is the coefficient of determination on the test set (higher is
    better).  Regularization parameter was selected from 10 log-spaced values in
    $[10^{-4}, 10^4]$ by $5$-fold cross-validation. 
}
\label{figure:low_budget_reg3}
\end{figure*}

In this experiment, we demonstrate the ability of the proposed framework to
reach good regression performance with a small number of bases $k$. We compared:
\begin{itemize}[topsep=0pt,itemsep=-1ex,partopsep=1ex,parsep=1ex]
    \item[a)] Proposed with $\mathcal{K}=\mathcal{H}^3$ (with augmented features),
    \item[b)] Proposed with $\mathcal{K}=\mathcal{A}^3$ (with augmented features),
    \item[c)] Nystr\"{o}m method with $\mathcal{K}=\mathcal{P}^3_1$ and
\item[d)] Random Selection: choose $\bs{p}_1,\dots,\bs{p}_k$ uniformly at random
    from training set and use $\mathcal{K}=\mathcal{P}^3_1$.
    \end{itemize}
For a) and b) we used the lifted approach. For fair comparison in terms of model
size (number of floats used), we set $r=k/3$.  Results on the \textit{abalone},
\textit{cadata} and \textit{cpusmall} datasets are shown in Figure
\ref{figure:low_budget_reg3}.  We see that \textit{i)} the proposed framework
reaches the same performance as kernel ridge regression with much fewer bases
than other methods and \textit{ii)} $\mathcal{H}^3$ tends to outperform
$\mathcal{A}^3$ on these tasks.  Similar trends were observed when using
$\mathcal{K}=\mathcal{H}^2$ or $\mathcal{A}^2$.

\section{Discussion}
\label{sec:discussion}
    
\textbf{Ability to infer weights of unobserved interactions.} In our view, one
of the strengths of PNs and FMs is their ability to infer the weights
of unobserved feature interactions, unlike traditional kernel methods.  To see
why, recall that in kernel methods, predictions are computed by $\hat{y} =
\sum_{i=1}^n \alpha_i \mathcal{K}(\bs{x}_i, \bs{x})$. When
$\mathcal{K}=\mathcal{H}^m$ or $\mathcal{A}^m$, by
Lemma~\ref{lemma:tensor_kernel}, this is equivalent to $\hat{y} = \langle
\widetilde{\tensor{W}}, \symtensor{\bs{x}}{m} \rangle$ or $\langle
\widetilde{\tensor{W}}, \symtensor{\bs{x}}{m} \rangle_>$ if we set
$\widetilde{\tensor{W}} \coloneqq \sum_{i=1}^n \alpha_i
\symtensor{\bs{x}_i}{m}$.  Thus, in kernel methods, the weight associated with
$x_{j_1} \dots x_{j_m}$ can be written as a linear combination of the training
data's monomials:
\begin{equation}
    \widetilde{\tensor{W}}_{j_1,\dots,j_m} = \sum_{i=1}^n \alpha_i x_{j_1i} \dots x_{j_mi}.
\end{equation}
Assuming binary features, the weights of monomials that were never observed in
the training set are zero.  In contrast, in PNs and FMs, we have $\tensor{W} =
\sum_{s=1}^k \lambda_i \symtensor{\bs{p}_s}{m}$ and therefore the weight
associated with $x_{j_1} \dots x_{j_m}$ becomes
\begin{equation}
    \tensor{W}_{j_1,\dots,j_m} = \sum_{s=1}^k \lambda_s p_{j_1s} \dots p_{j_ms}.
\end{equation}
Because parameters are shared across monomials, PNs and FMs are able to
\textit{interpolate} the weights of monomials that were never observed in the
training set. This is the key property which makes it possible to use them
on recommender system tasks. In future work, we plan to apply
PNs and FMs to biological data, where this property should be very useful,
e.g., for inferring higher-order interactions between genes.

\textbf{ANOVA kernel vs. polynomial kernel.} One of the key properties of
the ANOVA kernel $\combikern{p}{x}{m}$ is multi-linearity w.r.t. elements
of $\bs{p}$ (Lemma \ref{lemma:multi_linearity}).  This is the key difference with
$\homogkern{p}{x}{m}$ which makes the direct optimization objective multi-convex
when $\mathcal{K}=\mathcal{A}^m$ (Theorem~\ref{theorem:direct_obj_multiconvex}).
However, because we need to ignore irrelevant monomials, computing the
kernel and its gradient is more challenging. Deriving efficient training
algorithms for arbitrary $m \in \mathbb{N}$ is an important future work.

In our experiments in Section \ref{sec:exp_fit_lambda}, we showed that fixing
$\bs{\lambda}=\bs{1}$ works relatively well when $\mathcal{K}=\mathcal{A}^2$. To
see intuitively why this is the case, note that fixing
$\bs{\lambda}=\bs{1}$ is equivalent to constraining the weight matrix $\bs{W}$
to be positive semidefinite, i.e., $\exists \bs{P}$ s.t. $\bs{W}=\decomp{P}{P}$.
Next, observe that we can rewrite the prediction function as
\begin{equation}
    \hat{y}_{\mathcal{A}^2}(\bs{x}; \bs{1}, \bs{P}) = \langle \decomp{P}{P},
    \symtensor{\bs{x}}{2} \rangle_> = \langle \mathcal{U}(\decomp{P}{P}),
    \symtensor{\bs{x}}{2} \rangle,
\end{equation}
where $\mathcal{U}(\bs{M})$ is a mask which sets diagonal and lower-diagonal
elements of $\bs{M}$ to zero. We therefore see that when using
$\mathcal{K}=\mathcal{A}^2$, we are learning a \textit{strictly
upper-triangular matrix}, parametrized by $\decomp{P}{P}$.  Importantly, the
matrix $\mathcal{U}(\decomp{P}{P})$ is not positive semidefinite.  This is what
gives the model some degree of freedom, even though  $\decomp{P}{P}$ is positive
semidefinite.  In contrast, when using $\mathcal{K}=\mathcal{H}^2$, if we fix
$\bs{\lambda}=\bs{1}$, then we have that
\begin{equation} 
    \hat{y}_{\mathcal{H}^2}(\bs{x}; \bs{1}, \bs{P}) = \langle
    \decomp{P}{P}, \symtensor{\bs{x}}{2} \rangle = \bs{x}^\tr \decomp{P}{P}
\bs{x} \ge 0 
\end{equation} and therefore the model is unable to predict
negative values.

Empirically, we showed in Section \ref{sec:exp_low_budget} that $\mathcal{H}^m$
outperforms $\mathcal{A}^m$ for low-budget non-linear regression. In contrast,
we showed in Section \ref{sec:exp_recsys} that $\mathcal{A}^m$ outperforms
$\mathcal{H}^m$ for recommender systems. The main difference between the two
experiments is the nature of the features used: continuous for the former and
binary for the latter. For binary features, squared features $x_1^2, \dots,
x_d^2$ are redundant with $x_1, \dots, x_d$ and are therefore not expected to
help improve accuracy. On the contrary, they might introduce bias towards
first-order features. We hypothesize that the ANOVA kernel is in general a
better choice for binary features, although this needs to be verified by more
experiments, for instance on natural language processing (NLP) tasks.

\textbf{Direct vs. lifted optimization.} The main advantage of direct
optimization is that we only need to estimate $\bs{\lambda} \in \mathbb{R}^k$
and $\bs{P} \in \mathbb{R}^{d \times k}$ and therefore the number of parameters
to estimate is independent of the degree $m$. Unfortunately, the approach is
neither convex nor multi-convex when using $\mathcal{K}=\mathcal{H}^m$. In
addition, the regularized objective \eqref{eq:direct_regul_obj} is non-smooth
w.r.t. $\bs{\lambda}$. In Section \ref{sec:lifted_optimization}, we proposed to
reformulate the problem as one of low-rank symmetric tensor estimation and used
a symmetrization trick to obtain a multi-convex smooth objective function.
Because this objective involves the estimation of $m$ matrices of size $d \times
r$, we need to set $r = k / m$ for fair comparison with the direct objective in
terms of model size.  When $\mathcal{K}=\mathcal{A}^m$, we showed that the
direct objective is readily multi-convex. However, an advantage of our lifted
objective when $\mathcal{K}=\mathcal{A}^m$ is that it is convex w.r.t. larger
block of variables than the direct objective.  

\vspace{-0.1cm}
\section{Conclusion}

In this paper, we revisited polynomial networks \citep{livni} and factorization
machines \citep{fm,libfm} from a unified perspective. We proposed direct and
lifted optimization approaches and showed their equivalence in the regularized
case for $m=2$.  With respect to PNs, we proposed the first CD solver with
support for arbitrary integer $m \ge 2$. With respect to FMs, we made several
novel contributions including making a connection with the ANOVA kernel, proving
important properties of the objective function and deriving the first CD solver
for third-order FMs.  Empirically, we showed that the proposed algorithms
achieve excellent performance on non-linear regression and recommender system
tasks.

\clearpage

\section*{Acknowledgments} 

This work was partially conducted as part of ``Research and Development on
Fundamental and Applied Technologies for Social Big Data'', commissioned by the
National Institute of Information and Communications Technology (NICT), Japan.
We also thank Vlad Niculae, Olivier Grisel, Fabian Pedregosa and Joseph Salmon
for their valuable comments.

\bibliography{paper_icml}
\bibliographystyle{icml2016}

\clearpage
\onecolumn
\appendix

\begin{center}
    {\Huge \bf Supplementary material}
\end{center}

\section{Symmetric tensors}

\subsection{Background}

Let $\mathbb{R}^{d_1 \times \dots \times d_m}$ be the set of $d_1 \times \dots
\times d_m$ real $m$-order tensors. In this paper, we focus on cubical tensors,
i.e., $d_1 = \dots = d_m = d$. We denote the set of $m$-order cubical tensors by
$\mathbb{R}^{d^m}$. We denote the elements of $\tensor{M} \in \mathbb{R}^{d^m}$
by $\tensor{M}_{j_1,\dots,j_m}$, where $j_1,\dots,j_m \in [d]$.

Let $\bs{\sigma} = [\sigma_1,\dots,\sigma_m]$ be a permutation of
$\{1,\dots,m\}$. Given $\tensor{M} \in \mathbb{R}^{d^m}$, we define
$\tensor{M}_{\bs{\sigma}} \in \mathbb{R}^{d^m}$ as the tensor such that
\begin{equation}
    (\tensor{M}_{\bs{\sigma}})_{j_1,\dots,j_m} \coloneqq
    \tensor{M}_{j_{\sigma_1},\dots,j_{\sigma_m}} \quad \forall j_1,\dots,j_m \in
    [d].
\end{equation}
In other words $\tensor{M}_{\bs{\sigma}}$ is a copy of $\tensor{M}$ with its
axes permuted. This generalizes the concept of transpose to tensors.

Let $P_m$ be the set of all permutations of $\{1,\dots,m\}$. We say that a
tensor $\tensor{X} \in \mathbb{R}^{d^m}$ is symmetric if and only if
\begin{equation}
    \tensor{X}_{\bs{\sigma}} = \tensor{X} \quad \forall \bs{\sigma} \in P_m.
\end{equation}
We denote the set of symmetric tensors by $\mathbb{S}^{d^m}$.

Given $\tensor{M} \in \mathbb{R}^{d^m}$, we define the symmetrization of
$\tensor{M}$ by 
\begin{equation}
    \symop{\tensor{M}} = \frac{1}{m!} \sum_{\bs{\sigma} \in
        P_m} \tensor{M}_{\bs{\sigma}}.
\label{eq:symmetrized_tensor}
\end{equation}
Note that when $m=2$, then $\symop{\bs{M}} = \frac{1}{2} (\bs{M} + \bs{M}^\tr)$.

\begin{figure*}[t]
    \center
    \includegraphics[scale=0.38]{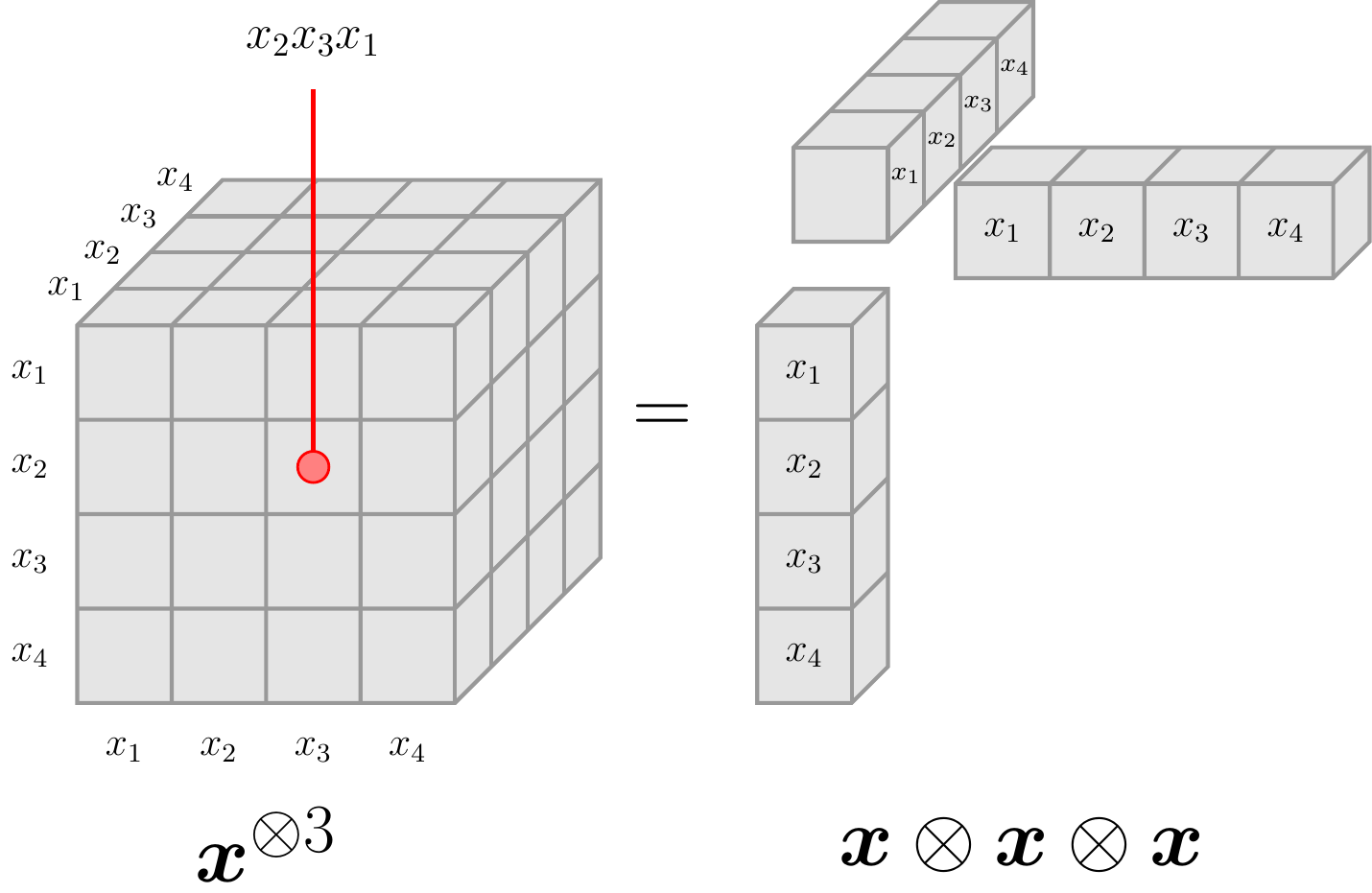}
    \hspace{1cm}
    \includegraphics[scale=0.38]{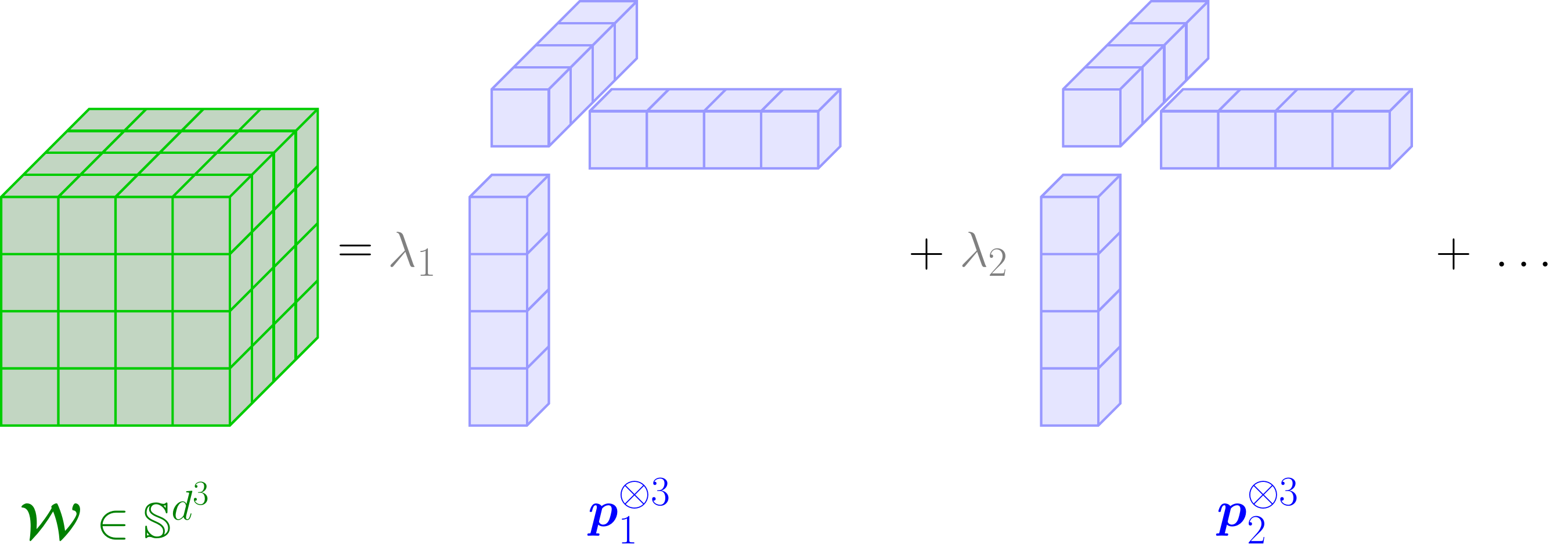}
\caption{Illustration of symmetric rank-one tensor (left) and symmetric outer
product decomposition (right).}
\label{figure:tensor_illust}
\end{figure*}

Given $\bs{x} \in \mathbb{R}^d$, we define a
symmetric rank-one tensor by $\symtensor{\bs{x}}{m} \coloneqq \underbrace{\bs{x} \otimes
\dots \otimes \bs{x}}_{m \text{ times}} \in \mathbb{S}^{d^m}$, i.e.,
$(\symtensor{\bs{x}}{m})_{j_1,j_2,\dots,j_m} = x_{j_1} x_{j_2} \dots x_{j_m}$.
We denote the symmetric outer product decomposition \citep{sym_tensors} of
$\tensor{W} \in \mathbb{S}^{d^m}$ by
\begin{equation}
\tensor{W} = \sum_{s=1}^k \lambda_s \symtensor{\bs{p}_s}{m},
\end{equation}
where $k$ is called the symmetric rank of $\tensor{W}$.
This generalizes the concept of eigendecomposition to tensors.
These two concepts are illustrated in Figure \ref{figure:tensor_illust}.

\subsection{Proof of Lemma \ref{lemma:sym_forms}}
\label{appendix:proof_sym_forms}

Assume $\tensor{M} \in
\mathbb{R}^{d^m}$ and $\tensor{X} \in \mathbb{S}^{d^m}$. Then,
\begin{align}
\langle \symop{\tensor{M}}, \tensor{X} \rangle 
&= \frac{1}{m!} \sum_{\bs{\sigma} \in
P_m} \langle \tensor{M}_{\bs{\sigma}}, \tensor{X} \rangle
\quad \quad &&\text{by definition of } \symop{\tensor{M}} \text{ and by linearity} \\
&= \frac{1}{m!} \sum_{\bs{\sigma} \in
P_m} \langle (\tensor{M}_{\bs{\sigma}})_{\bs{\sigma}^{-1}},
\tensor{X}_{\bs{\sigma}^{-1}} \rangle
\quad \quad &&\text{since } \langle \tensor{A}, \tensor{B} \rangle = \langle
\tensor{A}_{\bs{\sigma}}, \tensor{B}_{\bs{\sigma}} \rangle ~ \forall \tensor{A},
\tensor{B} \in \mathbb{R}^{d^m}, \forall \bs{\sigma} \in P_m \\
&= \frac{1}{m!} \sum_{\bs{\sigma} \in
P_m} \langle \tensor{M},
\tensor{X}_{\bs{\sigma}^{-1}} \rangle
\quad \quad &&\text{by definition of inverse permutation } \\
&= \frac{1}{m!} \sum_{\bs{\sigma} \in
P_m} \langle \tensor{M},
\tensor{X} \rangle
\quad \quad &&\text{since } \tensor{X} \in \mathbb{S}^{d^m} \\
            &= \langle \tensor{M}, \tensor{X} \rangle.
\end{align}

\section{Proofs related to ANOVA kernels}

\subsection{Proof of multi-linearity (Lemma \ref{lemma:multi_linearity})}
\label{appendix:proof_multi_linearity}

For $m=1$, we have
\begin{align}
\combikern{p}{x}{1} 
&= \sum_{j=1}^d p_j x_j \\
&= \sum_{k \neq j} p_k x_k + p_j x_j \\
&= \mathcal{A}^1(\negvec{p}{j}, \negvec{x}{j})
+ ~ p_j x_j ~ \mathcal{A}^{0}(\bs{p}_{\neg j}, \bs{x}_{\neg j}) 
\end{align}
where we used $\combikern{p}{x}{0} = 1$.

For $1 < m \le d$, first notice that we can rewrite
\eqref{eq:combination_kernel} as
\begin{align}
\combikern{p}{x}{m} &= 
\sum_{j_m > \dots > j_1} p_{j_1} x_{j_1} \dots p_{j_m} x_{j_m} 
\quad j_k \in [d], k \in [m]\\
&= \sum_{j_1=1}^{d-m+1} \sum_{j_2=j_1+1}^{d-m+2} \dots
    \sum_{j_m=j_{m-1}+1}^d p_{j_1} x_{j_1} \dots p_{j_m} x_{j_m}.
\end{align}
Then,
\begin{align}
\combikern{p}{x}{m} 
=& \sum_{j_1=1}^{d-m+1} \sum_{j_2=j_1+1}^{d-m+2} \dots \sum_{j_m=j_{m-1}+1}^d
p_{j_1} x_{j_1} p_{j_2} x_{j_2} \dots p_{j_m} x_{j_m} \\
=& \sum_{j_2=j_1+1}^{d-m+2} \dots \sum_{j_m=j_{m-1}+1}^d
p_1 x_1 p_{j_2} x_{j_2} \dots p_{j_m} x_{j_m} + \\
&\sum_{j_1=2}^{d-m+1} \sum_{j_2=j_1+1}^{d-m+2} \dots \sum_{j_m=j_{m-1}+1}^d
p_{j_1} x_{j_1} p_{j_2} x_{j_2} \dots p_{j_m} x_{j_m} \\
=& ~ p_1 x_1 \mathcal{A}^{m-1}(\negvec{p}{1}, \negvec{x}{1}) + 
\mathcal{A}^{m}(\negvec{p}{1}, \negvec{x}{1}).
\end{align}
We can always permute the elements of $\bs{p}$ and $\bs{x}$ without changing
$\combikern{p}{x}{m}$. It follows that
\begin{equation}
\combikern{p}{x}{m} =
p_j x_j \mathcal{A}^{m-1}(\negvec{p}{j}, \negvec{x}{j}) + 
\mathcal{A}^{m}(\negvec{p}{j}, \negvec{x}{j}) \quad \forall j \in [d].
\end{equation}

\subsection{Efficient computation when $m \in \{2,3\}$}
\label{appendix:linear_time_imperative}

Using the multinomial theorem, we can expand the homogeneous polynomial kernel
as
\begin{equation}
\homogkern{p}{x}{m} = \innerprod{p}{x}^m = \sum_{k_1 + \dots + k_d=m} \binom{m}{k_1, \dots, k_d} 
    \prod_{j=1}^d (p_j x_j)^{k_j}
\label{eq:multinomial_theorem}
\end{equation}
where
\begin{equation}
    \binom{m}{k_1, \dots, k_d} \coloneqq \frac{m!}{k_1! \dots k_d!}
\end{equation}
is the multinomial coefficient and $k_j \in \{0, 1, \dots, m\}$. Intuitively,
$\binom{m}{k_1, \dots, k_d}$ is the weight of the monomial $(p_1 x_1)^{k_1}
\dots (p_d x_d)^{k_d}$ in the expansion.  For instance, if $\bs{p}, \bs{x} \in
\mathbb{R}^3$, then the weight of $p_1 x_1 p_3^2 x_3^2$ is $\binom{3}{1, 0, 2} =
3$.  The main observation is that monomials where all $k_1, \dots, k_d$ are in
$\{0,1\}$ correspond to monomials of \eqref{eq:combination_kernel}. If we can
compute all other monomials efficiently, then we just need to subtract
them from the homogeneous kernel in order to obtain
\eqref{eq:combination_kernel}.

To simplify notation, we define the shorthands
\begin{equation}
    \rho_j \coloneqq p_j x_j, 
    \quad 
    \diagkern{p}{x}{m} \coloneqq \sum_{j=1}^d \rho_j^m
    \quad \text{and} \quad
    \diagkern{p}{x}{m,n} \coloneqq \diagkern{p}{x}{m} \diagkern{p}{x}{n}.
\end{equation}

\underline{Case $m=2$}

For $m=2$, the possible monomials are of the form $\rho_j^2$ for all $j$ and
$\rho_i \rho_j$ for $j > i$.  Applying \eqref{eq:multinomial_theorem}, we obtain
\begin{equation}
\begin{aligned}
\homogkern{p}{x}{2}
&= \sum_{j=1}^d \rho_j^2 + 2 \sum_{j > i} \rho_i \rho_j \\
&= \diagkern{p}{x}{2} + 2 \combikern{p}{x}{2}
\end{aligned}
\end{equation}
and therefore
\begin{equation}
    \combikern{p}{x}{2} = \frac{1}{2} \left[\homogkern{p}{x}{2} -
\diagkern{p}{x}{2}\right].
\end{equation}

This formula was already mentioned in \citep{stitson}.  It was also rediscovered
in \citep{fm,libfm}, although the connection with the ANOVA kernel was not
identified.

\underline{Case $m=3$}

For $m=3$, the possible monomials are of the form $\rho_j^3$ for all $j$,
$\rho_i \rho_j^2$ for $i \neq j$ and $\rho_i \rho_j \rho_k$ for $k > j > i$.
Applying \eqref{eq:multinomial_theorem}, we obtain
\begin{equation}
\begin{aligned}
\homogkern{p}{x}{3}
&= \sum_{j=1}^d \rho_j^3 + 3 \sum_{i \neq j} \rho_i
\rho_j^2 + 6 \sum_{k > i > i} \rho_i \rho_j \rho_k \\
&= \diagkern{p}{x}{3} + 3 \sum_{i \neq j} \rho_i
\rho_j^2 + 6 \combikern{p}{x}{3}.
\end{aligned}
\end{equation}
We can compute the second term efficiently by using
\begin{equation}
\begin{aligned}
\sum_{i \neq j} \rho_i \rho_j^2 
&= \sum_{i,j=1}^d \rho_i \rho_j^2 - \sum_{j=1}^d \rho_j^3 \\
&= \diagkern{p}{x}{2,1} - \diagkern{p}{x}{3}.
\end{aligned}
\end{equation}
We therefore obtain
\begin{equation}
\begin{aligned}
\combikern{p}{x}{3}
&= \frac{1}{6} \left[\homogkern{p}{x}{3} - \diagkern{p}{x}{3} - 3 \left(
\diagkern{p}{x}{2,1} - \diagkern{p}{x}{3}\right) \right] \\
&= \frac{1}{6} \left[\homogkern{p}{x}{3}  - 3 \diagkern{p}{x}{2,1} + 2
\diagkern{p}{x}{3} \right].
\end{aligned}
\end{equation}

\subsection{Proof of multi-convexity (Theorem
\ref{theorem:direct_obj_multiconvex})}
\label{appendix:theorem_direct_obj_multiconvex}

Let us denote the rows of $\bs{P}$ by $\col{p}_1, \dots, \col{p}_d \in
\mathbb{R}^k$.  Using Lemma \ref{lemma:multi_linearity}, we know that there
exists constants $a_s$ and $b_s$ such that for all $j \in [d]$
\begin{align}
    \hat{y}_{\mathcal{A}^m}(\bs{x};\bs{\lambda}, \bs{P}) 
    &= \sum_{s=1}^k \lambda_s \mathcal{A}^m(\bs{p}_s, \bs{x}) \\
&= \sum_{s=1}^k \lambda_s (p_{js} x_j a_s + b_s) \\
&= \sum_{s=1}^k p_{js} \lambda_s x_j a_s +
\text{const} \\
&= \langle \col{p}_j, \col{\mu}_j \rangle + \text{const}
\quad \text{where} \quad
\col{\mu}_j \coloneqq [\lambda_1 x_j a_1, \dots, \lambda_k x_j a_k]^\tr.
\end{align}
Hence $\hat{y}_{\mathcal{A}^m}(\bs{x};\bs{\lambda}, \bs{P})$ is an affine
function of $\col{p}_1, \dots, \col{p}_d$.  The composition of a convex loss
function and an affine function is convex.  Therefore, \eqref{eq:direct_obj} is
convex in $\col{p}_j ~ \forall j \in [d]$. Convexity w.r.t. $\bs{\lambda}$ is
obvious.

\section{Proof of equivalence between regularized problems
(Theorem \ref{theorem:equivalence_reg_problems})}
\label{appendix:proof_equivalence_reg_problems}

First, we are going to prove that the optimal solution of the nuclear norm
penalized problem is a symmetric matrix. For that, we need the following lemma.

\begin{mylemma}{Upper-bound on nuclear norm of symmetrized matrix}

\begin{equation}
    \|\symop{\bs{M}}\|_* \le \|\bs{M}\|_* \quad \forall \bs{M} \in
    \mathbb{R}^{d^2}
\end{equation}
\label{lemma:nuclear_norm_sym_matrix}
\end{mylemma}
\textit{Proof.}
\begin{equation}
\begin{aligned}
    \|\mathcal{\symop{\bs{M}}}\|_* 
    &=  \|\frac{1}{2}(\bs{M} + \bs{M}^\tr)\|_* \\
&= \frac{1}{2} (\|\bs{M} + \bs{M}^\tr\|_*) \\
&\le \frac{1}{2} (\|\bs{M}\|_* + \|\bs{M}^\tr\|_*) \\
&= \|\bs{M}\|_*,
\label{eq:sym_inequality}
\end{aligned}
\end{equation}
with equality in the third line holding if and only if $\bs{M} = \bs{M}^\tr$.
The second and third lines use absolute homogeneity and subadditivity, two
properties that matrix norms satisfy. The last line uses the fact that
$\|\bs{M}\|_* = \|\bs{M}^\tr\|_*$.  \hfill $\qed$

\begin{mylemma}{Symmetry of optimal solution of nuclear norm penalized problem}

 \begin{equation}
     \argmin_{\bs{M} \in \mathbb{R}^{d^2}} \bar{L}_{\mathcal{K}}(\bs{M}) \coloneqq 
     L_{\mathcal{K}}(\symop{\bs{M}}) + \beta \|\bs{M}\|_* \in \mathbb{S}^{d^2}
 \end{equation}
\label{lemma:sym_optimal}
\end{mylemma}
\textit{Proof.} From any (possibly asymmetric) square matrix $\bs{A} \in \mathbb{R}^{d^2}$,
we can construct $\bs{M} = \symop{\bs{A}}$. We obviously have
$L_{\mathcal{K}}(\symop{\bs{A}}) = L_{\mathcal{K}}(\symop{\bs{M}})$. Combining
this with Lemma \ref{lemma:nuclear_norm_sym_matrix}, we have that
$\bar{L}_{\mathcal{K}}(\bs{M}) \le \bar{L}_{\mathcal{K}}(\bs{A})$.  Therefore we
can always achieve the smallest objective value by choosing a symmetric matrix.
\hfill $\qed$

Next, we recall the variational formulation of the nuclear norm based on the
SVD.

\begin{mylemma}{Variational formulation of nuclear norm based on SVD}

\begin{equation}
\|\bs{M}\|_* = \underset{\substack{\bs{U},\bs{V}\\\bs{M}=\bs{U}\bs{V}^\tr}}{\min} ~
    \frac{1}{2} (\|\bs{U}\|_F^2 + \|\bs{V}\|_F^2) \quad \forall \bs{M} \in
    \mathbb{R}^{d^2}
\label{eq:nuclear_variational}
\end{equation}
The minimum above is attained at 
$\|\bs{M}\|_* = \frac{1}{2} (\|\bs{U}\|_F^2 + \|\bs{V}\|_F^2)$,
where $\bs{U} \in \mathbb{R}^{d \times r}$ and $\bs{V} \in \mathbb{R}^{d \times
r}$, $r = \rank(\bs{M})$, are formed from the reduced SVD of
$\bs{M}$, i.e., $\bs{U} = \bs{A} \diag(\bs{\sigma})^{\frac{1}{2}}$ and $\bs{V} =
\bs{B} \diag(\bs{\sigma})^{\frac{1}{2}}$ where $\bs{M} = \bs{A}
\diag(\bs{\sigma}) \bs{B}^\tr$.
\label{lemma:nuclear_variational}
\end{mylemma}
For a proof, see for instance \citep[Section A.5]{mazumder}.

Now, we give a specialization of the above
for symmetric matrices, based on the eigendecomposition instead of SVD.

\begin{mylemma}{Variational formulation of nuclear norm based on
eigendecomposition}
    
\begin{equation}
    \|\bs{M}\|_* = \underset{\substack{\bs{\lambda}, \bs{P}\\
    \bs{M} = \eigen{P}{\lambda}}}{\min}
\sum_{s=1}^k |\lambda_s| ~ \|\bs{p}_s\|^2 \quad \forall \bs{M} \in
\mathbb{S}^{d^2},
\label{eq:nuclear_variational_sym}
\end{equation}
where $k = \rank(\bs{M})$. The minimum above is attained by the reduced
eigendecomposition $\bs{M}=\eigen{P}{\lambda}$ and $\|\bs{M}\|_* =
\|\bs{\lambda}\|_1$.
\label{lemma:nuclear_variational_sym}
\end{mylemma}
\textit{Proof.} Let $\bs{A} \diag(\bs{\sigma}) \bs{B}^\tr$ and
$\eigen{P}{\lambda}$ be the reduced SVD and eigendecomposition of $\bs{M} \in
\mathbb{S}^{d^2}$, respectively. The relation between the SVD and the
eigendecomposition is given by
\begin{align}
    \sigma_s &= |\lambda_s| \\ 
    \bs{a}_s &= \sign(\lambda_s) \bs{p}_s \\
    \bs{b}_s &= \bs{p}_s.
\end{align}
From Lemma \ref{lemma:nuclear_variational}, we therefore obtain
\begin{align}
    \bs{u}_s &= \sqrt{\sigma_s} \bs{a}_s = \sqrt{|\lambda_s|} \sign(\lambda_s)
\bs{p}_s \\
\bs{v}_s &= \sqrt{\sigma_s} \bs{b}_s = \sqrt{|\lambda_s|}\bs{p}_s.
\end{align}
Now, computing $\frac{1}{2} (\sum_s \|\bs{u}_s\|^2 + \|\bs{v}_s\|^2)$ gives 
$\sum_{s=1}^k |\lambda_s| ~ \|\bs{p}_s\|^2$.
The minimum value $\|\bs{M}\|_* = \|\bs{\lambda}\|_1$ follows from the fact that
$\bs{P}$ is orthonormal and hence $\|\bs{p}_s\|^2=1 ~ \forall s \in [k]$. \hfill
$\qed$

We now have all the tools to prove our result.  The equivalence between
\eqref{eq:lifted_regul_obj} and \eqref{eq:nuclear_obj} when $r =
\rank(\bs{M}^*)$ is a special case of \citep[Theorem 3]{mazumder}.  From Lemma
\ref{lemma:sym_optimal}, we know that the optimal solution of
\eqref{eq:nuclear_obj} is symmetric. This allows us to substitute
\eqref{eq:nuclear_variational} with \eqref{eq:nuclear_variational_sym}, and
therefore, \eqref{eq:direct_regul_obj} is equivalent to \eqref{eq:nuclear_obj}
with $k = \rank(\bs{M}^*)$. As discussed in \citep{mazumder}, the result
also holds when $r=k$ is larger than $\rank(\bs{M}^*)$.

\section{Efficient coordinate descent algorithms}
\label{appendix:cd}

\subsection{Direct approach, $\mathcal{K}=\mathcal{A}^m$ for $m \in {\{2,3\}}$}
\label{appendix:direct_cd}

\begin{table}[t]
\caption{Examples of convex loss functions. We defined $\tau = \frac{1}{1
+ e^{-y \hat{y}}}$.}
\label{table_loss_func}
\begin{center}
\begin{footnotesize}
\begin{tabular}{c|c|c|c|c|c}
    Loss & Domain of $y$ & $\ell(y, \hat{y})$ & $\ell'(y, \hat{y})$ &
    $\ell''(y, \hat{y})$ & $\mu$ \\
\hline
    Squared & $\mathbb{R}$ & $\frac{1}{2} (\hat{y} - y)^2$ &
$\hat{y} - y$ & $1$ & $1$ \\
    Squared hinge & $\{-1, 1\}$ & $\max(1 - y \hat{y}, 0)^2$ &
$-2 y \max(1 - y \hat{y}, 0)$ & $2 \delta_{[\hat{y} y < 1]}$ & $2$ \\
Logistic & $\{-1, 1\}$ & $\log(\tau^{-1})$ &
$y (\tau - 1)$ & $\tau (1 - \tau)$ & $\frac{1}{4}$ \\
\hline
\end{tabular}
\end{footnotesize}
\end{center}
\end{table}

As stated in Theorem \ref{theorem:direct_obj_multiconvex}, the direct
optimization objective is multi-convex when $\mathcal{K}=\mathcal{A}^m$. This
allows us to easily minimize the objective by solving a succession of
coordinate-wise convex problems. In this section, we develop an efficient
algorithm for minimizing \eqref{eq:direct_regul_obj} with $m \in \{2,3\}$. It is
easy to see that minimization w.r.t. $\bs{\lambda}$ can be reduced to a standard
$\ell_1$-regularized convex objective via a simple change of variable. We
therefore focus our attention to minimization w.r.t. $\bs{P}$.

As a reminder, we want to minimize
\begin{equation}
    f \coloneqq \sum_{i=1}^n \ell(y_i, \hat{y}_i) + \beta \sum_{s=1}^k |\lambda_s|
    \|\bs{p}_s\|^2 
\end{equation}
where
\begin{equation}
    \hat{y}_i \coloneqq \sum_{s=1}^k \lambda_s \mathcal{A}^m(\bs{p}_s,
    \bs{x}_i).
\end{equation}
After routine calculation, we obtain
\begin{align}
    \partialfrac{\mathcal{A}^2(\bs{p}_s, \bs{x}_i)}{p_{js}} 
    &= \langle \bs{p}_s, \bs{x}_i \rangle x_{ji} - p_{js} x_{ji}^2 
    = (\langle \bs{p}_s, \bs{x}_i \rangle - p_{js} x_{ji}) x_{ji} \\
    \partialfrac{\mathcal{A}^3(\bs{p}_s, \bs{x}_i)}{p_{js}} 
    &= \frac{1}{2} \langle \bs{p}_s, \bs{x}_i \rangle^2 x_{ji} - p_{js} x_{ji}^2 
    \langle \bs{p}_s, \bs{x}_i \rangle 
    - \frac{1}{2} x_{ji} \mathcal{D}^2(\bs{p}_s, \bs{x}_i) + p_{js}^2 x_{ji}^3  \\
    &= \mathcal{A}^2(\bs{p}_s, \bs{x}_i) x_{ji} - p_{js} x_{ji}^2 
    \langle \bs{p}_s, \bs{x}_i \rangle + p_{js}^2 x_{ji}^3  \\
    \partialfrac{\mathcal{A}^m(\bs{p}_s, \bs{x}_i)}{p_{js}^2} 
    &= 0 \quad \forall m \in \mathbb{N}\\
    \partialfrac{\hat{y}_i}{p_{js}} &= \lambda_s
    \partialfrac{\mathcal{A}^m(\bs{p}_s, \bs{x}_i)}{p_{js}} \\
    \partialfrac{\hat{y}_i}{p_{js}^2} &= 0 \quad \forall j \in [d], s \in [k].
\end{align}
The fact that the second derivative is null is a consequence of the
multi-linearity of $\mathcal{A}^m$.

Using the chain rule, we then obtain
\begin{align}
    \partialfrac{f}{p_{js}} &= \sum_{i=1}^n \ell'(y_i, \hat{y}_i)
    \partialfrac{\hat{y}_i}{p_{js}} + 2 \beta |\lambda_s| p_{js} \\
    \partialfrac{f}{p_{js}^2} 
    &= \sum_{i=1}^n \left[\ell''(y_i, \hat{y}_i)
\left(\partialfrac{\hat{y}_i}{p_{js}}\right)^2 + \ell'(\hat{y}_i,y_i)
\partialfrac{\hat{y}_i}{p_{js}^2} \right] + 2 \beta |\lambda_s| \\
    &= \sum_{i=1}^n \ell''(y_i, \hat{y}_i)
    \left(\partialfrac{\hat{y}_i}{p_{js}}\right)^2 + 2 \beta |\lambda_s|.
\end{align}
Assuming that $\ell$ is $\mu$-smooth, its second derivative is
upper-bounded by $\mu$ and therefore we have
\begin{equation}
    \partialfrac{f}{p_{js}^2} \le \eta_{js} \quad \text{where} \quad
    \eta_{js} \coloneqq \mu \sum_{i=1}^n
    \left(\partialfrac{\hat{y}_i}{p_{js}}\right)^2 + 2 \beta |\lambda_s|.
\end{equation}
Then the update 
\begin{equation}
    p_{js} \leftarrow p_{js} - \eta_{js}^{-1} \partialfrac{f}{p_{js}}
\end{equation}
guarantees that the objective value is monotonically decreasing except at the
coordinate-wise minimum. Note that in the case of the squared loss
$\ell(y,\hat{y})=\frac{1}{2}(y - \hat{y})^2$, the above update is equivalent to
a Newton step and is the exact minimizer of the coordinate-wise objective.  An
epoch consists in updating all variables once, for instance in cyclic order.

For an efficient implementation, we need to maintain $\hat{y}_i ~ \forall i \in
[n]$ and statistics that depend on $\bs{p}_s$.  For the former, we need $O(n)$
memory. For the latter, we need $O(kmn)$ memory for an implementation with
full cache. However, this requirement is not realistic for a large training set.
In practice, the memory requirement can be reduced to $O(mn)$ if we recompute
the quantities then sweep through $p_{1s}, \dots, p_{ds}$ for $s$ fixed.
Overall the cost of one epoch is $O(k n_z(\bs{X}))$.  A similar implementation
technique is described for factorization machines with $m=2$ in \cite{libfm}. 

\subsection{Lifted approach, $\mathcal{K}=\mathcal{H}^m$}
\label{appendix:lifted_cd}

\begin{figure*}[t]
\begin{minipage}[t]{0.48\linewidth} 
\begin{algorithm}[H]
    \caption{\footnotesize CD algorithm for direct obj. with
    $\mathcal{K}=\mathcal{A}^{\{2,3\}}$}
\begin{algorithmic}
\footnotesize
\STATE {\bfseries Input:} $\bs{\lambda}$, initial $\bs{P}$, $\mu$-smooth loss
function $\ell$, regularization parameter $\beta$, number of bases $k$,
degree $m$, tolerance $\epsilon$
\STATE Pre-compute $\hat{y}_i \coloneqq \hat{y}_{\mathcal{A}^m}(\bs{x}_i;
\bs{\lambda}, \bs{P}) ~ \forall i \in  [n]$
\STATE Set $\Delta \leftarrow 0$
\FOR{$s \coloneqq 1, \dots, k$}
\STATE Pre-compute $\langle \bs{p}_s, \bs{x}_i \rangle$ and $\mathcal{A}^2(\bs{p}_s, \bs{x}_i) ~ \forall i \in [n]$
        \FOR{$j \coloneqq 1, \dots, d$}
            \STATE Compute inv. step size $\eta \coloneqq \mu \sum_{i=1}^n
            \left(\partialfrac{\hat{y}_i}{p_{js}}\right)^2 + 2 \beta
            |\lambda_s|$
            \STATE Compute $\delta \coloneqq \eta^{-1} \left[\sum_{i=1}^n
            \ell'(y_i, \hat{y}_i) \partialfrac{\hat{y}_i}{p_{js}} + 2 \beta
        |\lambda_s| p_{js} \right]$
            \STATE Update $p_{js} \leftarrow p_{js} - \delta$;
            Set $\Delta \leftarrow \Delta + |\delta|$
                \STATE Synchronize $\hat{y}_i$,
$\langle \bs{p}_s, \bs{x}_i \rangle$ and
                $\mathcal{A}^2(\bs{p}_s, \bs{x}_i)$ $\forall i$
                s.t.  $x_{ji} \neq 0$
        \ENDFOR
\ENDFOR
\STATE If $\Delta \le \epsilon$ stop, otherwise repeat
\STATE {\bfseries Output:} $\bs{P}$
\end{algorithmic}
\label{algo:direct_cd}
\end{algorithm}
\end{minipage}
\begin{minipage}[t]{0.48\linewidth} 
\begin{algorithm}[H]
    \caption{\footnotesize CD algorithm for lifted objective with
    $\mathcal{K}=\mathcal{H}^m$}
\begin{algorithmic}
\footnotesize
\STATE {\bfseries Input:} initial $\{\bs{U}^t\}_{t=1}^m$, $\mu$-smooth loss
function $\ell$, regularization parameter $\beta$, rank $r$, 
degree $m$, tolerance $\epsilon$
\STATE Pre-compute $\hat{y}_i \coloneqq \sum_{s=1}^r \prod_{t=1}^m \langle
\bs{u}_s^t, \bs{x}_i \rangle ~ \forall i \in  [n]$
\STATE Set $\Delta \leftarrow 0$
\FOR{$t \coloneqq 1, \dots, m$ and $s \coloneqq 1, \dots, r$}
    \STATE Pre-compute $\xi_i \coloneqq \prod_{t' \neq
    t} \langle \bs{u}^{t'}_s, \bs{x}_i \rangle ~ \forall i \in [n]$
        \FOR{$j \coloneqq 1, \dots, d$}
            \STATE Compute inv. step size $\eta \coloneqq \mu \sum_{i=1}^n
            \xi_i^2 x_{ji}^2 + \beta$
            \STATE Compute $\delta \coloneqq \eta^{-1} \left[\sum_{i=1}^n
            \ell'(y_i, \hat{y}_i) \xi_i x_{ji} + \beta u^t_{js}\right]$
            \STATE Update $u^t_{js} \leftarrow u^t_{js} - \delta$; 
            Set $\Delta \leftarrow \Delta + |\delta|$
            \STATE Synchronize $\hat{y}_i$ $\forall i$ s.t.
            $x_{ji} \neq 0$ 
        \ENDFOR
\ENDFOR
\STATE If $\Delta \le \epsilon$ stop, otherwise repeat
\STATE {\bfseries Output:} $\{\bs{U}^t\}_{t=1}^m$
\end{algorithmic}
\label{algo:lifted_cd}
\end{algorithm}
\end{minipage}
\end{figure*}

We present an efficient coordinate descent solver for the lifted approach with
$\mathcal{K}=\mathcal{H}^m$, for arbitrary integer $m \ge 2$. Recall
that our goal is to learn $\tensor{W} = \symop{\tensor{M}} \in \mathbb{S}^{d^m}$
by factorizing $\tensor{M} \in \mathbb{R}^{d^m}$ using $m$ matrices of size $d
\times r$. Let us call these matrices $\bs{U}^1, \dots, \bs{U}^m$ and their
columns $\bs{u}^t_s = [u^t_{1s}, \dots, u^t_{ds}]^\tr$ with $t \in [m]$ and $s
\in [r]$. The decomposition of $\tensor{M}$ can be expressed as a sum of
rank-one tensors
\begin{equation}
    \tensor{M} = \sum_{s=1}^r \bs{u}^1_s \otimes \dots \otimes \bs{u}^m_s.
\end{equation}

Using \eqref{eq:sym_forms} we obtain
\begin{align}
    \hat{y}_i \coloneqq \langle \tensor{W}, \symtensor{\bs{x}_i}{m} \rangle  =
\langle \tensor{M}, \symtensor{\bs{x}_i}{m} \rangle 
              = \sum_{s=1}^r \prod_{t=1}^m \langle \bs{u}_s^t, \bs{x}_i \rangle.
\end{align}
The first and second coordinate-wise derivatives are given by
\begin{align}
    \partialfrac{\hat{y}_i}{u_{js}^t} = \prod_{t' \neq t} \langle \bs{u}^{t'}_s,
    \bs{x}_i \rangle
    x_{ji} \quad \text{and} \quad
    \partialfrac{\hat{y}_i}{(u_{js}^t)^2} = 0.
\end{align}
We consider the following regularized objective function
\begin{equation}
    f \coloneqq \sum_{i=1}^n \ell(y_i, \hat{y}_i) + \frac{\beta}{2} \sum_{t=1}^m
    \sum_{s=1}^r \|\bs{u}^t_s\|^2.
\end{equation}
Using the chain rule, we obtain
\begin{align}
\partialfrac{f}{u_{js}^t} = \sum_{i=1}^n \ell'(y_i, \hat{y}_i)
\partialfrac{\hat{y}_i}{u_{js}^t} + \beta u^t_{js} 
\quad \text{and} \quad
\partialfrac{f}{(u_{js}^t)^2} = \sum_{i=1}^n \ell''(y_i, \hat{y}_i)
\left(\partialfrac{\hat{y}_i}{u_{js}^t}\right)^2 + \beta.
\end{align}
Assuming that $\ell$ is $\mu$-smooth, its second derivative is
upper-bounded by $\mu$ and therefore we have
\begin{equation}
    \partialfrac{f}{(u_{js}^t)^2} \le \eta^t_{js} \quad \text{where} \quad
    \eta^t_{js} \coloneqq \mu \sum_{i=1}^n
    \left(\partialfrac{\hat{y}_i}{u_{js}^t}\right)^2 + \beta.
\end{equation}
Then the update 
\begin{equation}
    u^t_{js} \leftarrow u^t_{js} - (\eta^t_{js})^{-1} \partialfrac{f}{u_{js}^t}
\end{equation}
guarantees that the objective value is monotonically decreasing, except at the
coordinate-wise minimum. Note that in the case of the squared loss
$\ell(y,\hat{y})=\frac{1}{2}(y - \hat{y})^2$, the above update is equivalent to
a Newton step and is the exact minimizer of the coordinate-wise objective. An
epoch consists in updating all variables once, for instance in cyclic order.

For an efficient implementation, the two quantities we need to maintain are
$\hat{y}_i ~ \forall i \in [n]$ and $\prod_{t' \neq t} \langle \bs{u}^{t'}_s,
\bs{x}_i \rangle ~ \forall i \in [n], s \in [r], t \in [m]$.  For the former, we
need $O(n)$ memory. For the latter, we need $O(rmn)$ memory for an implementation
with full cache. However, this requirement is not realistic for a large training
set. In practice, the memory requirement can be reduced to $O(mn)$ if we
recompute the quantity then sweep through $u^t_{1s}, \dots, u^t_{ds}$ for $t$
and $s$ fixed.  Overall the cost of one epoch is $O(mr n_z(\bs{X}))$.  

\subsection{Lifted approach, $\mathcal{K}=\mathcal{A}^2$}
\label{appendix:lifted_cd_anova}

For $\langle \cdot, \cdot \rangle_>$, efficient computations are more involved
since we need to ignore irrelevant monomials. Nevertheless, we can also compute
the predictions directly without explicitly symmetrizing the model. For
$m=2$, it suffices to subtract the effect of squared features. It is easy to
verify that we then obtain
\begin{equation}
\footnotesize
\langle \symop{\decomp{U}{V}}, \symtensor{\bs{x}}{2} \rangle_>
= \frac{1}{2} \left[
\langle \bs{U}^\tr \bs{x}, \bs{V}^\tr \bs{x} \rangle - 
\sum_{s=1}^r \langle \bs{u}_s \circ \bs{x}, \bs{v}_s \circ \bs{x} \rangle
\right],
\end{equation}
where $\circ$ indicates element-wise product. The coordinate-wise derivatives
are given by
\begin{equation}
    \partialfrac{y_i}{u_{js}} = \frac{1}{2} \left[ \langle \bs{v}_s, \bs{x} \rangle
x_{ji} - v_{js} x_{ji}^2 \right] \quad \text{and} \quad
\partialfrac{y_i}{v_{js}} = \frac{1}{2} \left[ \langle \bs{u}_s, \bs{x} \rangle
x_{ji} - u_{js} x_{ji}^2 \right].
\end{equation}
Generalizing this to arbitrary $m$ is a future work.

\section{Datasets}
\label{appendix:datasets}

For regression experiments, we used the following public datasets.
\begin{table}[H]
\label{table:datasets_reg}
\begin{center}
    \begin{tabular}{c|c|c|c|c}
    Dataset & $n$ (train) & $n$ (test) & $d$ & Description\\
\hline
abalone & 3,132 & 1,045 & 8 & Predict the age of abalones from physical measurements \\
cadata & 15,480 & 5,160 & 8 & Predict housing prices from economic covariates \\
cpusmall & 6,144 & 2,048 & 12 & Predict a computer system activity from system performance measures \\
diabetes & 331 & 111 & 10 & Predict disease progression from baseline
measurements \\
E2006-tfidf & 16,087 & 3,308 & 150,360 & Predict volatility of stock returns
from company financial reports \\
\hline
\end{tabular}
\end{center}
\end{table}

The \textit{diabetes} dataset is available in scikit-learn \citep{sklearn}.
Other datasets are available from
\url{http://www.csie.ntu.edu.tw/~cjlin/libsvmtools/datasets/}.

For recommender system experiments, we used the following two public datasets.
\begin{table}[H]
\label{table:datasets_recsys}
\begin{center}
\begin{tabular}{c|c|c}
Dataset & $n$ & $d$ \\
\hline
Movielens 1M & 1,000,209 (ratings) & 9,940 = 6,040 (users) + 3,900 (movies) \\
Last.fm & 108,437 (tag counts) & 24,078 = 12,133 (artists) + 11,945 (tags) \\
\hline
\end{tabular}
\end{center}
\end{table}

For \textit{Movielens 1M}, the task is to predict ratings between 1 and 5 given
by users to movies, i.e., $y \in \{1, \dots, 5\}$. For \textit{Last.fm}, the
task is to predict the number of times a tag was assigned to an artist, i.e., $y
\in \mathbb{N}$.  

The design matrix $\bs{X}$ was constructed following \citep{fm,libfm}. Namely,
for each rating $y_i$, the corresponding $\bs{x}_i$ is set to the concatenation
of the one-hot encodings of the user and item indices. Hence the number of
samples $n$ is the number of ratings and the number of features is equal to the
sum of the number of users and items. Each sample contains exactly two non-zero
features. It is known that factorization machines are equivalent to matrix
factorization when using this representation \citep{fm,libfm}.

We split samples uniformly at random between 75\% for training and
25\% for testing. 

\end{document}